\documentclass{article}

\usepackage{microtype}
\usepackage{graphicx}
\usepackage{subfigure}
\usepackage{booktabs} 
\usepackage{wrapfig}
\usepackage{bm}
\usepackage{multirow}
\usepackage{array}
\usepackage{colortbl}
\usepackage{dashrule}
\usepackage{arydshln}
\usepackage{capt-of} 
\usepackage{amsmath}

\usepackage{hyperref}

\usepackage[accepted]{icml2025}

\usepackage{amsmath}
\usepackage{amssymb}
\usepackage{mathtools}
\usepackage{amsthm}

\usepackage[capitalize,noabbrev]{cleveref}

\theoremstyle{plain}

\theoremstyle{definition}

\theoremstyle{remark}

\usepackage[textsize=tiny]{todonotes}

\icmltitlerunning{VTGaussian-SLAM: RGBD SLAM for Large Scale Scenes with Splatting View-Tied 3D Gaussians}

\begin{document}

\twocolumn[
\icmltitle{VTGaussian-SLAM: RGBD SLAM for Large Scale Scenes with Splatting View-Tied 3D Gaussians}




\begin{icmlauthorlist}
\icmlauthor{Pengchong Hu}{yyy}
\icmlauthor{Zhizhong Han}{yyy}
\end{icmlauthorlist}

\icmlaffiliation{yyy}{Machine Perception Lab, Wayne State University, Detroit, USA}

\icmlcorrespondingauthor{Zhizhong Han}{h312h@wayne.edu}

\icmlkeywords{Machine Learning, ICML}

\vskip 0.3in
]



\printAffiliationsAndNotice{This project was partially supported by an NVIDIA academic award and a Richard Barber research award.} 

\begin{abstract}
Jointly estimating camera poses and mapping scenes from RGBD images is a fundamental task in simultaneous localization and mapping (SLAM). State-of-the-art methods employ 3D Gaussians to represent a scene, and render these Gaussians through splatting 
for higher efficiency and better rendering.
However, these methods cannot scale up to extremely large scenes, 
due to the inefficient tracking and mapping strategies that need to optimize all 3D Gaussians in the limited GPU memories throughout the training to maintain the geometry and color consistency to previous RGBD observations. To resolve this issue, we propose novel tracking and mapping strategies to work with a novel 3D representation, dubbed view-tied 3D Gaussians, for RGBD SLAM systems. View-tied 3D Gaussians is a kind of simplified Gaussians, 
which is tied to depth pixels, without needing to learn locations, rotations, and multi-dimensional variances. Tying Gaussians to views not only significantly saves storage but also allows us to employ many more Gaussians to represent local details in the limited GPU memory. 
Moreover, our strategies remove the need of maintaining all Gaussians learnable throughout the training, 
while improving rendering quality, and tracking accuracy. We justify the effectiveness of these designs, and report better performance over the latest methods on the widely used benchmarks in terms of rendering and tracking accuracy and scalability. Please see our project page for code and videos at \href{https://machineperceptionlab.github.io/VTGaussian-SLAM-Project}{https://machineperceptionlab.github.io/VTGauss-ian-SLAM-Project}.
\end{abstract}

\section{Introduction}
SLAM is an important task in computer vision~\cite{keetha2024splatam,Zhu2021NICESLAM}, 3D reconstruction~\cite{wang2023coslam,Hu2023LNI-ADFP}, and robotics~\cite{9712211robot}. SLAM methods resolve the computational problem of 
mapping unknown environments while tracking camera locations. Traditional SLAM methods usually use RGB or RGBD images to sense 3D surroundings and track spatial relationships among different frames. The resulting 3D maps are usually represented by 3D point clouds, which are discrete and not friendly to downstream applications that require continuous geometry representations, such as geometry modeling, editing, and virtual reality.

Due to the continuity and the ability to represent arbitrary topology and appearance, neural representations have been employed with differentiable rendering in SLAM systems~\cite{zhang2023goslam,tofslam,teigen2023rgbd,sandström2023uncleslam,sucar2021imap,wang2023coslam}. Previous methods~\cite{zhang2023goslam,tofslam,teigen2023rgbd,Hu2023LNI-ADFP} learn neural radiance fields (NeRF)~\cite{mildenhall2020nerf}, a continuous implicit function 
modeling scene geometry and appearance, enabling image rendering for scene mapping and camera tracking.
More recently, 3DGS~\cite{kerbl3Dgaussians} was proposed for high-quality and real-time rendering, which also provides a novel perspective for time-sensitive SLAM problems~\cite{MatsukiCVPR2024_monogs,hhuang2024photoslam,yan2023gs,yugay2023gaussianslam,sandstrom2024splatslam}. 3DGS can efficiently render 3D Gaussians that model radiance fields explicitly using differentiable rasterization which is faster than the ray tracing in NeRF. 
However, maintaining numerous 3D Gaussians to cover the whole scene while ensuring color and geometry consistency across previous frames often leads to poor rendering in tracking and mapping. 
This obstacle makes 3DGS still hard to scale up to extremely large scenes in SLAM, remaining the challenge of improving the rendering quality, tracking accuracy, and scalability of 3DGS in tracking cameras and mapping scenes.

\begin{figure*}[t]
  \centering
   \includegraphics[width=\linewidth]{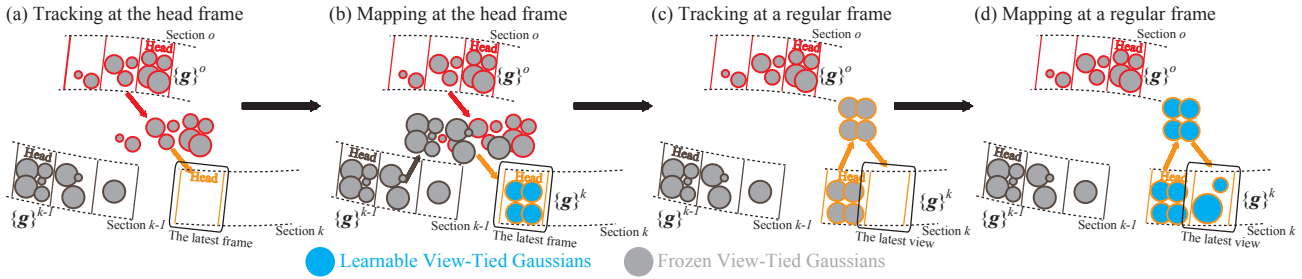}
   \vskip -0.1in
   \caption{Overview. (a) and (c) are tracking strategies, while (b) and (d) are mapping strategies. Please refer to Sec. 3.1 for more details.}
   \label{fig:overview}
\vskip -0.1in
\end{figure*}

To overcome this challenge, we propose an RGBD SLAM system with splatting view-tied 3D Gaussians. Our method introduces a novel point-based volume representation, dubbed view-tied 3D Gaussians, to represent the color and geometry of the scene, dedicated to reducing storage but pursuing better rendering quality. Different from the original 3D Gaussians~\cite{kerbl3Dgaussians}, we tie a 3D Gaussian to each pixel on the depth, which makes their positions only determined by depth and camera poses, without a need of learning and storing their locations and also the density control. Meanwhile, we simplify an ellipsoid 3D Gaussian as a sphere, without saving rotations and multi-dimensional variances as well, which saves more storage for initiating even more Gaussians for describing more details. With view-tied 3D Gaussians, our novel tracking and mapping strategies can be conducted in rendering and optimizing the Gaussians 
that are merely related to the most recent views, rather than all Gaussians in the scene. 
These benefits enable us to use many more Gaussians to represent each frame, just keep the most relevant Gaussians in the limited GPU memories, and remove the need of maintaining the consistency of scene representations to keyframes. This ability scales up the size of scenes that we can handle and also significantly improves the rendering quality, even if just using simplified Gaussians. Our visual and numerical evaluations show our advantages over the latest methods. Our main contributions are listed below.


\begin{itemize}
\item We propose view-tied Gaussian splatting that significantly reduces storage but improves rendering quality with 3DGS in SLAM.
\item We introduce a novel RGBD SLAM algorithm with view-tied Gaussian splatting. Our tracking and mapping strategies remove the need of holding and optimizing all Gaussians in memory throughout the training, which improves the scalability of 3DGS in SLAM.
\item We report the state-of-the-art results on widely used benchmarks over the latest 3DGS-based SLAM.
\end{itemize}

\section{Related Work}

\noindent\textbf{Multi-view Reconstruction. }Recent multi-view reconstruction methods aim to learn neural implicit representations~\cite{park2021nerfies,mueller2022instant,ruckert2021adop,yu_and_fridovichkeil2021plenoxels,Zhu2021NICESLAM,NeuralRGBDSurfaceReconstruction_2022_CVPR,wang2022go-surf,bozic2021transformerfusion,DBLP:journals/corr/abs-2209-15153,sun2021neucon,li2023rico} from multi-view images through volume rendering. Besides RGB images, we can also leverage depth~\cite{Yu2022MonoSDF,NeuralRGBDSurfaceReconstruction_2022_CVPR,Zhu2021NICESLAM,zhang2024volumerender} and normals~\cite{Yu2022MonoSDF,wang2022neuris,guo2022manhattan,patel2024normalguideddetailpreservingneuralimplicit} as rendering supervision to infer more geometry details. With the emergence of 3DGS~\cite{kerbl3Dgaussians,3dgrt2024}, we can learn implicit representations or accurate depth maps 
for reconstruction~\cite{zhang2024gspull,Huang2DGS2024,Yu2024GOF,wolf2024gs2mesh,fan2024trim,gslrm2024,charatan23pixelsplat,lin2024directlearningmeshappearance,chen2024pgsrplanarbasedgaussiansplatting}. However, these methods require accurate camera poses, which differs from SLAM methods a lot.

\noindent\textbf{SLAM with Volume Rendering. }With multi-view consistency, multi-view stereo (MVS)~\cite{schoenberger2016sfm,schoenberger2016mvs} estimate dense depth maps from multiple RGB images. 
With the demand for novel view synthesis, recent SLAM methods render depth maps and estimate continuous implicit representations~\cite{zhang2023goslam,tofslam,teigen2023rgbd,sandström2023uncleslam,sucar2021imap,wang2023coslam,Sandström2023ICCVpointslam,sandström2023uncleslam,cp-slam}. They use RGBD images as rendering supervision, and learn implicit representations through learning radiance fields.
Some other methods also leverage priors like segmentation priors~\cite{kong2023vmap,haghighi2023neural}, depth fusion priors~\cite{Hu2023LNI-ADFP}, or object-level priors~\cite{kong2023vmap}, which improve the performance in tracking and mapping. 

\noindent\textbf{SLAM with 3D Gaussian Splatting. }Because of the rendering efficiency and higher rendering quality, more recent methods used 3DGS to differentially render images~\cite{keetha2024splatam,MatsukiCVPR2024_monogs,hhuang2024photoslam,yan2023gs,yugay2023gaussianslam,sandstrom2024splatslam}. Although these methods adopt different tracking and mapping strategies, they need to maintain all Gaussians covering the scene in the limited GPU memories and optimize all Gaussians throughout the training to keep color and geometry consistency to all previous views. This fact limits the number of Gaussians that they can use and makes it hard to scale up to extremely large scenes. Unlike these methods, our method employed simplified Gaussians, without storing locations, rotations, or multi-view variances, saving more room for more Gaussians to represent either more details or larger scenes. Our view-tied Gaussians combine the advantages of Gaussian representations in current SLAM systems~\cite{keetha2024splatam,yugay2023gaussianslam}.

More complicated systems~\cite{liso2024loopyslam,zhu2024_loopsplat,bruns2024neuralgraphmapping,sandstrom2024splatslam} also leverage loop closure in the optimization. However, detecting loop closures among views needs pretrained priors and is sensitive to image quality, which also differs from ours.

\noindent\textbf{Gaussian Alignment. }Aligning 3D Gaussians to some entities was also employed in some other works~\cite{xu2024grm,gslrm2024,gao2024meshbasedgaussiansplattingrealtime,luiten2023dynamic3dgs,seidenschwarz2024dynomoonlinepointtracking,zakharov2024gh}. Gaussians in these methods are either not related to camera positions or with many attributes. Instead, our view-tied Gaussians relate Gaussians' positions with the camera poses and use simplified attributes, dedicated to improving the efficiency and scalability of SLAM systems.

\section{Methods}
\subsection{Overview}
Fig.~\ref{fig:overview} illustrates our tracking and mapping strategies. We organize view-tied 3D Gaussians from several consecutive frames as a section so that we can keep as many Gaussians as the GPU memory allows to represent a local area, access these Gaussians more efficiently, and more importantly, 
enable more robust completion of missing depth by utilizing neighboring frames' depth.
In each section, we mark the first frame as a head to differentiate it from 
regular frames for different Gaussian initialization strategies in mapping.   

With view-tied Gaussians, we manage to keep learnable Gaussians that are the most relevant to the latest frame in the GPU memory. This ability removes the need of maintaining a list of keyframes and optimizing all Gaussians representing the scene to maintain their spatial and appearance consistency to all keyframes, which are widely employed in the latest SLAM systems~\cite{wang2023coslam,Hu2023LNI-ADFP,keetha2024splatam,Zhu2021NICESLAM}, leading to the key of scaling up to much larger scenes. 

For tracking the latest frame, we select Gaussians in a section, 
render them from the camera pose initialized by the constant speed assumption, and optimize the pose by minimizing rendering errors to the latest frame. 
If the latest frame is the head of a new section in Fig.~\ref{fig:overview} (a), we select the Gaussians in a certain section in front according to the visibility. These selected Gaussians maintain the spatial consistency in a long view sequence and reduce the error cumulation. If the latest frame is not a head but a regular frame in the current section in Fig.~\ref{fig:overview} (c), we select the Gaussians in this section for renderings with higher quality.

\begin{figure}[t]
  \centering
\vskip 0.1in
   \includegraphics[width=\linewidth]{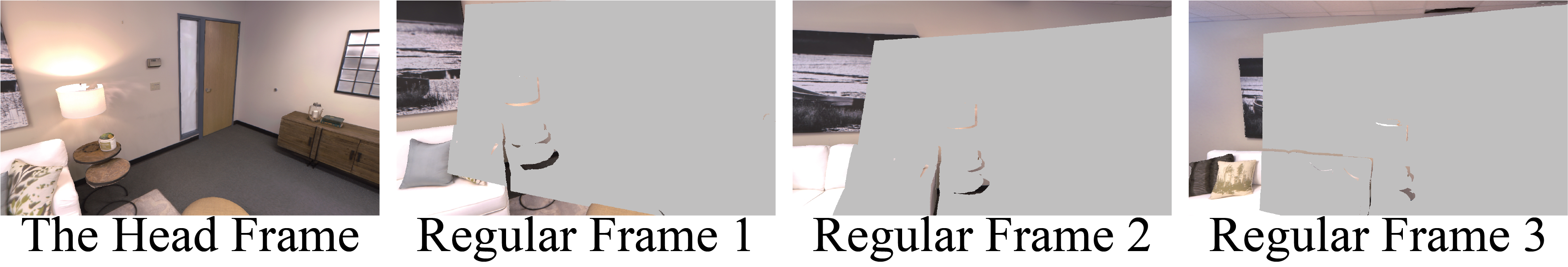}
   \vskip -0.1in
   \caption{Initialization of view-tied Gaussians in a section.}
   \label{fig:completion}
\vskip -0.1in
\end{figure}

For mapping the scene using the latest frame, if the latest frame is the head of a new section in Fig.~\ref{fig:overview} (b), we initialize Gaussians by centering them at all pixels with valid depth values. If the latest frame is not a head but a regular frame in an existing section in Fig.~\ref{fig:overview} (d), we only initialize Gaussians as a complement in areas where pixels have valid depth values and the existing Gaussians in the current section cannot cover. Fig.~\ref{fig:completion} visualizes the view-tied Gaussians initialized at both the head frame and the following regular frames in a section. We learn these Gaussians to maintain their color and geometry consistency to all frames in the same section and some previous frames with overlaps.

\subsection{View-tied Gaussians}
Our view-tied Gaussians aim to achieve memory efficiency in SLAM, which enables us to improve the rendering quality by using many more Gaussians to represent local details. Specifically, we simplify an ellipsoid Gaussian $\bm{g}$ into a sphere, which only includes a color $\bm{c}\in\mathbb{R}^{1\times3}$, a radius as the variance $r\in\mathbb{R}^1$, and an opacity $o\in\mathbb{R}^1$, and removes the 4-dimensional rotation, the 3-dimensional location, the other 2-dimensional variances from the original 3D Gaussians, saving $64.3\%$ (9/14) storage in total.

We remove the need of learning and storing locations by tying a Gaussian $\bm{g}$ at each pixel with a valid depth value on the depth map. We center $\bm{g}$ at the 3D location back-projected from the depth value. Thus, the positions of Gaussians are only determined by the depth and the corresponding camera pose, which can be adjusted in the camera tracking procedure. For the $i$-th frame with a RGB $\bm{V}_i$ and a depth map $\bm{D}_i$, we will initialize Gaussians $\{\bm{g}_j^i$\} on $\bm{D}_i$.

Since there may be missing depth values on some depth maps, we organize Gaussians from every $N$ neighboring frames into a section $\bm{S}_k$, so that we can use Gaussians from neighboring frames to cover the missing area on one specific depth map in the same section. Based on the concept of sections, we only optimize Gaussians in one section containing the latest view, and frozen Gaussians in other sections. This not only enables us to use more Gaussians 
to represent local details, but also removes the need to maintain the appearance and geometry consistency to all the previous frames or keyframes. This is a key to improving our rendering quality, leading to better performance in tracking and mapping, and scaling up to larger scenes.


\begin{figure*}[t]
  \centering
   \includegraphics[width=\linewidth]{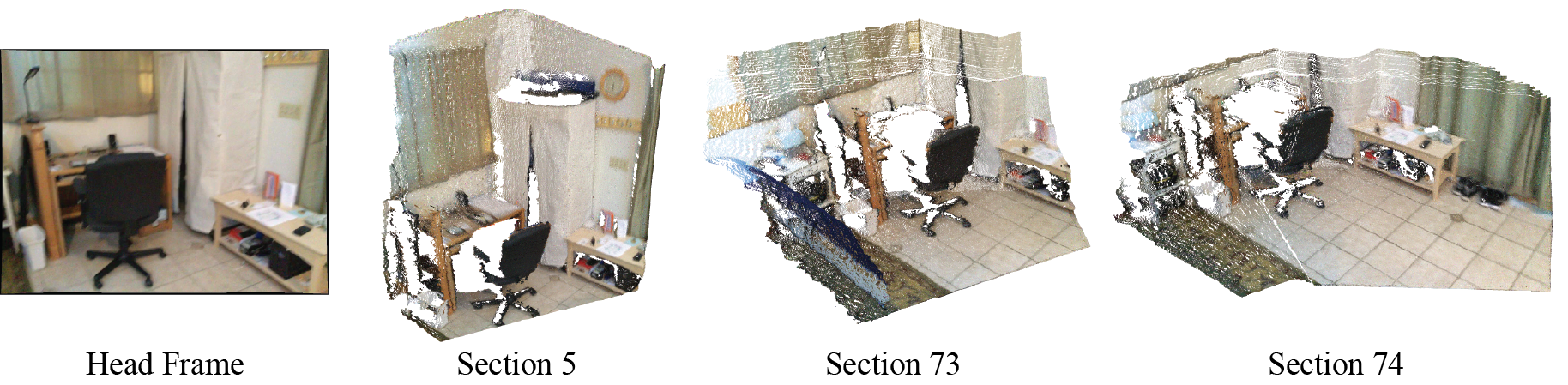}
   \vskip -0.1in
   \caption{Illustration of selecting overlapping section. We show Gaussian centers and colors in each section. }
   \label{fig:overlapping}
\vskip -0.1in
\end{figure*}

\subsection{Tracking Cameras}
We will estimate the camera pose $\bm{p}_i$ of the latest frame $\{\bm{V}_i,\bm{D}_i\}$ first. When tracking cameras, we keep all Gaussians in the scene fixed, and merely optimize the pose $\bm{p}_i$ by minimizing rendering errors with respect to $\{\bm{V}_i,\bm{D}_i\}$. 

If the latest frame $\{\bm{V}_i,\bm{D}_i\}$ is a head starting a new section $\bm{S}_k$, we select one section $\bm{S^o}$ in front of $\bm{S}_k$, and render Gaussians in $\bm{S^o}$ into a RGB $\bm{V}_i'$ and a depth $\bm{D}_i'$ using the pose $\bm{p}_i$. We optimize $\bm{p}_i$ to minimize rendering errors,
\begin{equation}
\label{Eq:trackingrendering}
\min_{\bm{p}_i} \ \alpha \bm{W}_i ||\bm{V}_i-\bm{V}_i'||_1 + \beta \bm{W}_i ||\bm{D}_i-\bm{D}_i'||_1,
\end{equation}

\noindent where $\{\bm{V}_i',\bm{D}_i'\}=splat(\{\bm{g}\}^o\in\bm{S^o},\bm{p}_i)$ are rendered images by splatting $\{\bm{g}\}^o$ in section $\bm{S^o}$ using the camera pose $\bm{p}_i$, $\bm{W}_i$ is a mask which removes pixels either without depth values or uncovered by the rendering of $\{\bm{g}\}^o$ or invisible to sections $\bm{S^o}$, and $\alpha$ and $\beta$ are balance parameters.

Otherwise, if the latest frame $\{\bm{V}_i,\bm{D}_i\}$ is a regular frame in the current section $\bm{S}_k$, we will optimize the camera pose $\bm{p}_i$ using the same equation above but rendering the Gaussians $\{\bm{g}\}^k$ in the current section $\bm{S}_k$ into $\{\bm{V}_i',\bm{D}_i'\}$.

\begin{wrapfigure}{r}{0.63\linewidth}
\centering
\vskip -0.2in
\includegraphics[width=\linewidth]{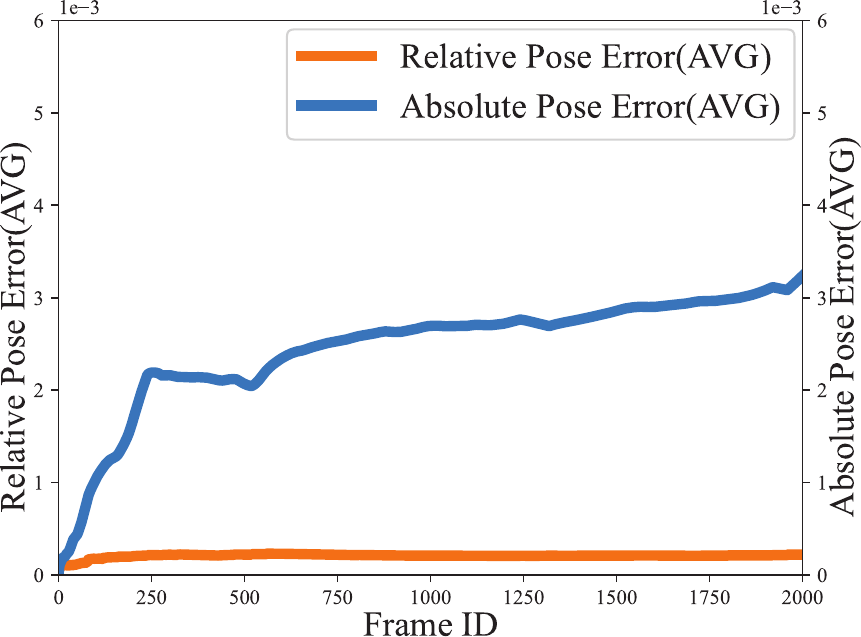}
\vskip -0.1in
    \caption{Issue of pose error cumulation.}
    \label{fig:poseissue}
 \vskip -0.05in
\end{wrapfigure} 

This design aims to find a balance between the rendering quality and the spatial consistency of the current section to previous frames in a long image sequence. Obviously, rendering Gaussians in the same section will produce better renderings since neighboring frames usually have larger overlaps with the latest frame. Although better renderings are helpful for more accurate camera pose estimations, the higher accuracy is merely meaningful relative to the neighboring frames, resulting in a significant cumulation of pose errors relative to the whole camera trajectory. We illustrate this issue in Fig.~\ref{fig:poseissue}, where we use Gaussians in the current section and render pretty good images for tracking. At each frame out of a $2000$ frame video, the average error of relative pose to the previous frame is pretty small, while the average error relative to the whole trajectory is getting larger and larger. To resolve this issue, we maintain the spatial consistency of the current section by rendering Gaussians $\{\bm{g}\}^o$ in a certain previous section $\bm{S^o}$ for tracking the head frame while rendering Gaussians $\{\bm{g}\}^k$ in the same section $\bm{S}_k$ for tracking regular frames.


\noindent\textbf{Selecting Overlapping Section $\bm{S}^o$. }In each section, we select a section $\bm{S}^o$ for the head frame in terms of overlap and visibility. We first set up an overlap candidate view list with an interval of $N_1$ frames. For a head frame $\{\bm{V}_i,\bm{D}_i\}$, we project the depth $\bm{D}_i$ from the initialized pose to each one frame in the overlap candidate view list. We count the number of visible points to each candidate view, and select the candidate views that have more visible points than a threshold $\gamma$. We eventually select the most front $3$ sections containing at least one of the selected candidate views. We use the most front as a criterion to relieve the impact caused by the error cumulation.

Then, we adopt a pre-tracking strategy to finalize the selection. We use each one of the $3$ sections to conduct the tracking for several iterations, respectively. We eventually select a section that produces the minimum rendering error as $\bm{S^o}$. Here, we do not splat Gaussians from all the $3$ sections together for rendering since multiple sections are seldom learned together in the mapping procedure, which may degenerate rendering quality. Fig.~\ref{fig:overlapping} illustrates the $3$ section candidates selected for a head frame, where all the $3$ sections are highly related to the head frame. We finally render Gaussians in section $73$ to track the head frame.

\noindent\textbf{Visibility Check. }We determine the visibility of a 3D point to a specific view if its projected depth is within $1\%$ of the interpolated depth at its projection on the depth map. For visibility to a section $\bm{S}^o$ with $\bm{W}_i$ in Eq.~\ref{Eq:trackingrendering}, we consider the head frame, the frame in the middle, and the last frame in $\bm{S}^o$ to calculate three visibility masks. We use the union of these three visibility masks as the visibility mask to a section. Please see Fig.~\ref{fig:visibility} for more details.

\subsection{Mapping Scenes}
We initialize Gaussians and learn attributes of Gaussians in the mapping procedure. In each section, we first initialize Gaussians at all pixels on the depth map at the head frame, and then complement Gaussians at pixels uncovered by the current Gaussians' renderings on regular frames. To maintain the appearance and geometry consistency to frames in both the same section and nearby overlapping sections, we adopt different strategies to learn Gaussian attributes.

If the latest frame $\{\bm{V}_i,\bm{D}_i\}$ is a head frame starting a new section $\bm{S}_k$, we use the Gaussians $\{\bm{g}\}^k$ initialized on the latest frame, the Gaussians $\{\bm{g}\}^{k-1}$ in the previous section $\bm{S}_{k-1}$, and the Gaussians $\{\bm{g}\}^o$ in the section $\bm{S}^o$ having the largest overlaps with the latest frame to render $\{\bm{V}_i',\bm{D}_i'\}=splat([\{\bm{g}\}^k,\{\bm{g}\}^{k-1},\{\bm{g}\}^{o}],\bm{p}_i)$. We minimize the rendering errors with respect to observations,

\vskip -0.25in
\begin{equation}
\label{Eq:mappingrendering1}
\min_{\{\bm{g}\}^k} \ \rho  ||\bm{V}_i-\bm{V}_i'||_1 + \tau L_{S}(\bm{V}_i,\bm{V}_i')+ \sigma \bm{U}_i ||\bm{D}_i-\bm{D}_i'||_1,
\end{equation}
\vskip -0.15in

\noindent where $L_{S}$ is the SSIM loss, $\bm{U}_i$ is a mask which removes pixels without valid depth values, $\rho$, $\tau$, $\sigma$ are balance weights. Only $\{\bm{g}\}^k$ are learnable, and all other Gaussians are fixed. This rendering can maintain the appearance and geometry consistency to the nearby overlapping sections.

\begin{figure}[t]
  \centering
   \includegraphics[width=\linewidth]{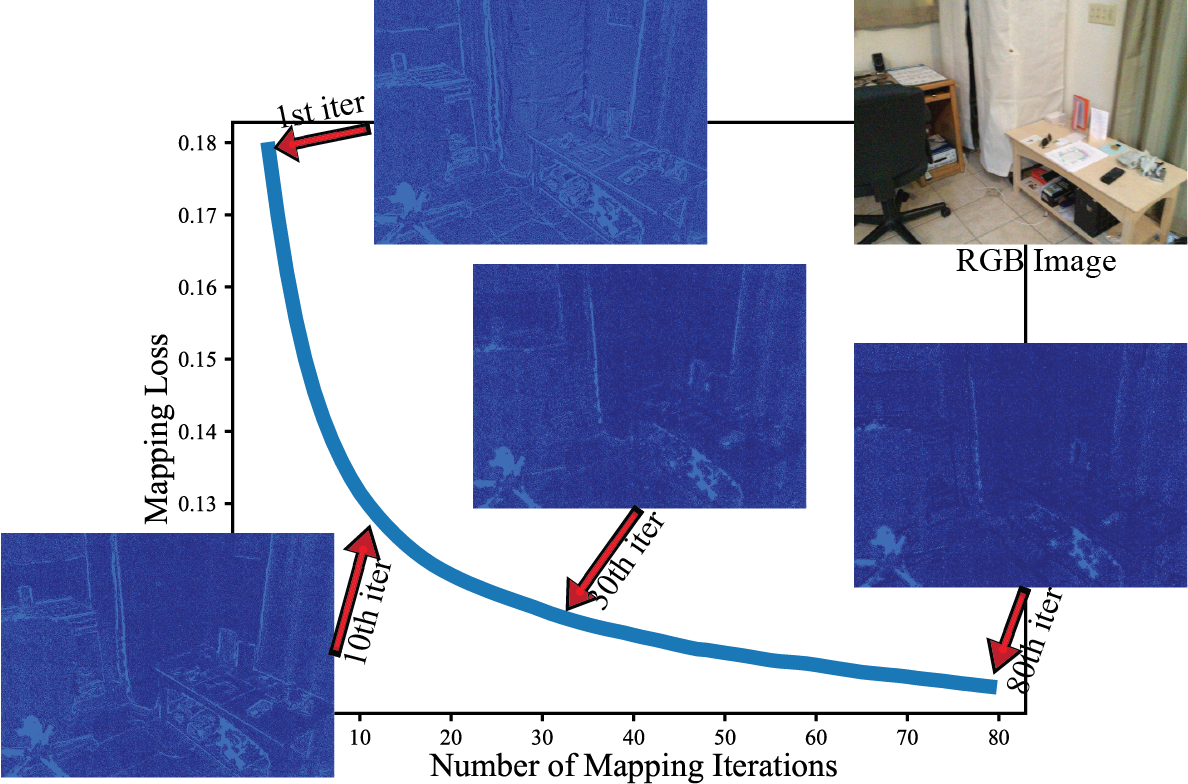}
   \vskip -0.05in
   \caption{Illustration of optimizing view-tied Gaussians initialized on a head frame. Error maps are shown at different iterations.}
   \label{fig:GaussianOptimization}
\vskip -0.2in
\end{figure}

Otherwise, if the latest frame $\{\bm{V}_i,\bm{D}_i\}$ is a regular frame in the current section $\bm{S}_k$, we first splat the Gaussians $\{\bm{g}\}^k$ in the current section into silhouette images to find the uncovered area where we initialize Gaussians. Then, we splat $\{\bm{g}\}^k$ to render from a random view $\bm{p}_j$ of $\{\bm{V}_j,\bm{D}_j\}$ that is in front of $\{\bm{V}_i,\bm{D}_i\}$ in this section, i.e., $\{\bm{V}_j',\bm{D}_j'\}=splat(\{\bm{g}\}^k,\bm{p}_j)$, where $\{\bm{V}_j,\bm{D}_j\}\in \bm{S}_k$ and $j\leq i$. We still use Eq.~\ref{Eq:mappingrendering1} to optimize $\{\bm{g}\}^k$, where we select a random view as a rendering target to maintain the appearance and geometry consistency to frames in the same section. We visualize the rendered images during mapping a head frame in Fig.~\ref{fig:GaussianOptimization}, where the rendering error is progressively minimized.

\begin{figure*}[!]
  \centering
   \includegraphics[width=\linewidth]{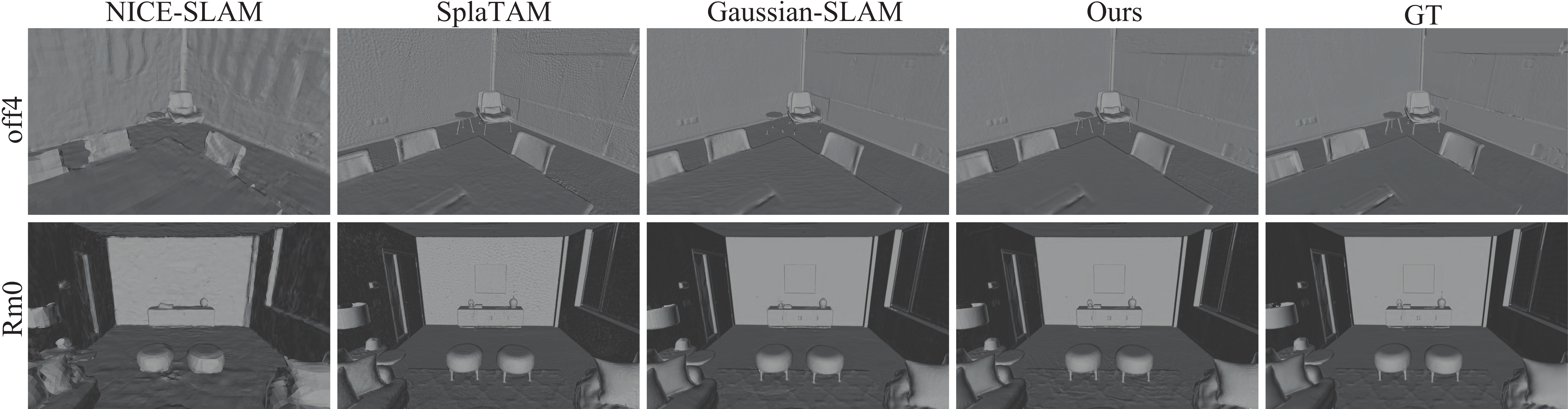}
   \vskip -0.1in
   \caption{Visual comparisons in reconstruction on Replica.}
   \label{fig:reconstructionReplica}
\vskip -0.15in
\end{figure*}

\begin{table*}
 \centering
\caption{Tracking comparisons in ATE RMSE $\downarrow [\mathrm{cm}]$ on Replica. $*$ denotes use of pre-trained data-driven priors.}
\vskip 0.1in

  \resizebox{\linewidth}{!}{
\begin{tabular}{ccccccc!{\vrule width 1pt}cccccc}
\toprule
 & \multicolumn{6}{c}{Neural Implicit Fields} & \multicolumn{6}{c}{3D Gaussian Splatting} \\
\midrule
 Methods & NICE-SLAM & DF-Prior & Vox-Fusion & ESLAM & Point-SLAM & Loopy-SLAM$*$ & SplaTAM & GS-SLAM & Gaussian-SLAM & LoopSplat$*$ & CG-SLAM$*$ & Ours \\
\midrule
 Avg. & 1.95 & 1.81 & 0.65 & 0.63 & 0.52 & 0.29 & 0.36 & 0.50 & 0.31 & 0.26 & 0.27 & \textbf{0.28} \\

\bottomrule
\end{tabular}}
\label{Tab:camreplicaperscene}
\vskip -0.1in
\end{table*}

\subsection{Bundle Adjustment}

We also conduct bundle adjustment to jointly estimate the camera pose and learn Gaussians when mapping the head frame in each section. We still use the Eq.~\ref{Eq:mappingrendering1} in the optimization, but also back-propagate gradients to update the camera pose of the head frame. We do not use bundle adjustment on each regular view in a section for stabilizing the optimization in both tracking and mapping.

\section{Experiments and Analysis}

We only report average results in this section, evaluations on each scene can be found in the supplementary materials. Please also watch our video for more renderings.

\noindent\textbf{Implementation Details.}
For neighboring views in a section $\bm{S}_k$, we choose $N=40$ on Replica~\cite{replica}, $N=30$ on TUM-RGBD~\cite{6385773ATERMSE_tumrgbd}, $N=30$ or $N=50$ on ScanNet~\cite{scannet}, and $N=100$ on ScanNet++~\cite{yeshwanthliu2023scannetpp}, according to image resolutions and mapping iterations.
During tracking, we use every 5 frames as a candidate view for overlapping section selection, \textit{i.e.} $N_1 = 5$. 
Specifically, when mapping a regular frame in an existing section $\bm{S}_k$, we will choose 0.5 as a mask threshold to determine if current Gaussians in $\bm{S}_k$ cover a pixel, if they do not, we will initialize a Gaussian as a complement.
To work with depth and RGB images with different quality and resolutions, we set $\rho=0.8$, $\tau=0.2$, and $\sigma=1.0$ to balance the mapping loss terms while we set $\{\alpha, \beta\} = \{0.5, 0.025\}$ on Replica, $\{0.5, 1.0\}$ on TUM-RGBD and ScanNet++, $\{0.5, 0.9\}$ on ScanNet to balance the tracking loss terms. More details of hyperparameters are provided in the supplementary materials.

\noindent\textbf{Dataset and Metrics.}
We conduct evaluations on several widely used benchmarks, including Replica~\cite{replica}, TUM-RGBD~\cite{6385773ATERMSE_tumrgbd}, ScanNet~\cite{scannet}, and ScanNet++~\cite{yeshwanthliu2023scannetpp}. Here Replica is a synthetic dataset with high-fidelity 3D reconstruction of indoor scenes. We evaluate on the widely used RGBD sequence from eight scenes captured by Sucar~\yrcite{sucar2021imap} with accurate trajectories. TUM-RGBD, ScanNet, and ScanNet++ are real-world datasets. 
Note that ScanNet++ is not a dataset designed for SLAM tasks, some sudden large motions are occurring in the DSLR-captured sequences, we follow previous methods~\cite{yugay2023gaussianslam,zhu2024_loopsplat} and only employ the first 250 frames of each scene in evaluations, which present smooth trajectories.

We evaluate the accuracy of estimated cameras at each frame and the rendering quality from either the observed or unobserved view angles. For tracking accuracy, we use the root mean square absolute trajectory error (ATE RMSE)~\cite{6385773ATERMSE_tumrgbd} as a metric. Regarding rendering quality, we measure PSNR, SSIM~\cite{1284395_ssim}, and LPIPS~\cite{zhang2018unreasonableeffectivenessdeepfeatures_lpips}. Similar to \cite{Sandström2023ICCVpointslam,liso2024loopyslam,zhu2024_loopsplat,yugay2023gaussianslam}, all the rendering metrics are computed by rendering the full resolution images along the estimated trajectory every $5$ frames. Additionally, we also obtain the meshes of scenes by marching cubes~\cite{Lorensen87marchingcubes} following a similar procedure in \cite{Sandström2023ICCVpointslam}. Then we measure the reconstruction performance with F1-score, the harmonic mean of the Precision (P) and Recall (R), using a distance threshold of 1 cm for all evaluations. We also use the depth L1 metric to measure the rendered mesh depth error at sampled novel views as in \cite{Zhu2021NICESLAM}.

\begin{figure}[t]
  \centering
   \includegraphics[width=\linewidth]{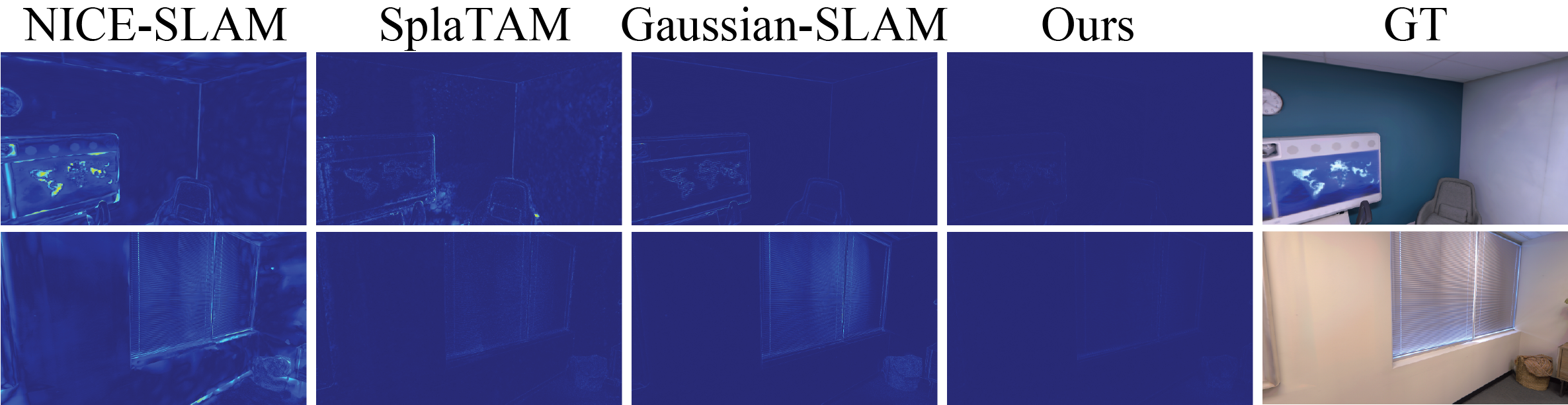}
   \vskip -0.1in
   \caption{Error map comparisons in rendering on Replica.}
   \label{fig:TackingReplica}
\vskip -0.15in
\end{figure}

\noindent\textbf{Baselines.}
We compare our method with the latest RGBD SLAM methods, including NeRF-based RGBD SLAM methods: NICE-SLAM~\cite{Zhu2021NICESLAM}, Vox-Fusion~\cite{Yang_Li_Zhai_Ming_Liu_Zhang_2022_voxfusion}, ESLAM~\cite{johari-et-al-2023-ESLAM}, DF-Prior~\cite{Hu2023LNI-ADFP}, Co-SLAM~\cite{wang2023coslam}, and Point-SLAM~\cite{Sandström2023ICCVpointslam}, and 3DGS-based RGBD SLAM methods: SplaTAM~\cite{keetha2024splatam}, MonoGS~\cite{MatsukiCVPR2024_monogs}, GS-SLAM~\cite{yan2023gs}, and Gaussian-SLAM~\cite{yugay2023gaussianslam}. Note that Point-SLAM~\cite{Sandström2023ICCVpointslam} requires ground truth depth
images as an input to guide sampling when rendering, which is an unfair advantage compared to other methods. Moreover, relying on data-driven priors, such as pre-trained NetVLAD models~\cite{Arandjelovic16}, in loop closure detection and visibility check, SLAM methods with pose graph optimizations Loopy-SLAM~\cite{liso2024loopyslam} and LoopSplat~\cite{zhu2024_loopsplat}, and CG-SLAM~\cite{hu2024cg} usually reported higher tracking accuracy, which however is not a fair experimental setting to most SLAM methods without using priors.

\subsection{Comparisons}

\begin{table}
    \caption{Rendering results in PSNR $\uparrow$, SSIM $\uparrow$, and LPIPS $\downarrow$ on three Datasets. $*$ denotes use of pre-trained data-driven priors.}
\vskip 0.1in
    \setlength{\tabcolsep}{2pt}
    \renewcommand{\arraystretch}{1.2}
	\centering
	\resizebox{\linewidth}{!}{
   \begin{tabular}{l|ccc|ccc|ccc}
    \toprule
    \textbf{Dataset} & \multicolumn{3}{c}{\textbf{\emph{Replica}}} & \multicolumn{3}{c}{\textbf{\emph{TUM}}} & \multicolumn{3}{c}{\textbf{\emph{ScanNet}}} \\
     \hline
      Method  &  PSNR $\uparrow$ &  SSIM $\uparrow$ &LPIPS $\downarrow$ &  PSNR $\uparrow$ &  SSIM $\uparrow$ &LPIPS $\downarrow$ &  PSNR $\uparrow$ &  SSIM $\uparrow$ &LPIPS $\downarrow$\\
    \hline
   \rowcolor{gray!20}
\multicolumn{10}{l}{\textit{Neural Implicit Fields}} \\
    NICE-SLAM &24.42& 0.809 & 0.233& 14.86& 0.614 & 0.441 & 17.54 & 0.621 & 0.548  \\
    Vox-Fusion &24.41 & 0.801 & 0.236 &16.46 & 0.677 &0.471 & 18.17 & 0.673 & 0.504 \\
   ESLAM & 28.06 & 0.923 &0.245 & 15.26 &0.478 & 0.569&  15.29& 0.658 &  0.488 \\
     Point-SLAM  & 35.17 &0.975  & 0.124&  16.62 &  0.696 &  0.526 & 19.82&  0.751& 0.514\\
     Loopy-SLAM$*$ &  35.47 &  0.981  &  0.109 & 12.94 &  0.489 & 0.645 &15.23 & 0.629 & 0.671 \\
     \hline
     \hline
\rowcolor{gray!20}
\multicolumn{10}{l}{\textit{3D Gaussian Splatting}} \\
    SplaTAM & 34.11 & 0.970 & 0.100 &  22.80 &  0.893& 0.178& 19.14 &  0.716 &  0.358 \\
    Gaussian-SLAM & 42.08 & \textbf{0.996} & 0.018 & 25.05& 0.929 & 0.168& 27.70 & 0.923 & 0.248 \\
    LoopSplat$*$ & 36.63 & 0.985 & 0.112 & 22.72 & 0.873 & 0.259 & 24.92 & 0.845 & 0.425 \\
    \hline
    \hline
   Ours & \textbf{43.34} & \textbf{0.996} & \textbf{0.012} & \textbf{30.20} & \textbf{0.972} & \textbf{0.062} & \textbf{31.10} & \textbf{0.961} & \textbf{0.108} \\
    \hline
    \end{tabular}}
    \vskip -0.15in
\label{Tab:rendering}
\end{table}



\begin{table*}[t]
\vskip -0.1in
\caption{Reconstruction results in D-L1 $[\mathrm{cm}]\downarrow$ and F1 $[\mathrm{\%}]\uparrow$ on Replica. $*$ denotes use of pre-trained data-driven priors.}
\vskip 0.1in
  \centering
  \resizebox{\linewidth}{!}{
\begin{tabular}{lcccccc!{\vrule width 1pt}ccccc}
\toprule
& \multicolumn{6}{c}{Neural Implicit Fields} & \multicolumn{5}{c}{3D Gaussian Splatting} \\
\midrule
 & NICE-SLAM & Vox-Fusion & ESLAM & Co-SLAM & Point-SLAM & Loopy-SLAM$*$ & SplaTAM & GS-SLAM & Gaussian-SLAM & LoopSplat$*$ & Ours \\
\midrule
D-L1$\downarrow$ & 2.97 & 2.46 & 1.18 & 2.59 & \textbf{0.44} & 0.35 & 0.72 & 1.16 & 0.68 &  0.51 & 0.53 \\
F1$\uparrow$ & 43.9 & 52.2 & 79.1 & 69.7 & 89.8 & 90.8 & 86.1 & 70.2 & 88.9 & 90.4 & \textbf{90.0} \\

\bottomrule
\end{tabular}}
\label{Tab:reconreplicaperscene}
\vskip -0.1in
\end{table*}

\begin{table*}[t]
\caption{Tracking comparisons in ATE RMSE $\downarrow[\mathrm{cm}]$ on TUM-RGBD. $*$ denotes use of pre-trained data-driven priors.}
\vskip 0.1in
  \centering
  \resizebox{\linewidth}{!}{
\begin{tabular}{ccccc!{\vrule width 1pt}cccccc}
\toprule
& \multicolumn{4}{c}{Neural Implicit Fields} & \multicolumn{6}{c}{3D Gaussian Splatting} \\
\midrule
 Methods & NICE-SLAM & Vox-Fusion & Point-SLAM & Loopy-SLAM$*$ & SplaTAM & GS-SLAM & Gaussian-SLAM & LoopSplat$*$ & CG-SLAM$*$ & Ours \\
\midrule
 Avg. & 13.3 & 10.3 & 3.0 & 2.9 & 3.3 & 3.7 & 2.9 & 2.3 & 2.0 & \textbf{2.6} \\

\bottomrule
\end{tabular}}
\vskip -0.1in

\label{Tab:camtumperscene}
\end{table*}

\noindent\textbf{Results on Replica. }We first report our results on 8 scenes in Replica. We compare with the state-of-the-art NeRF-based and GS-based SLAM methods in camera tracking in Tab.~\ref{Tab:camreplicaperscene}, mapping scenes with rendered images in Tab.~\ref{Tab:rendering}, and reconstruction in Tab.~\ref{Tab:reconreplicaperscene}.

Tab.~\ref{Tab:camreplicaperscene} shows that our method can estimate camera poses more accurately than state-of-the-art NeRF-based methods, such as NICE-SLAM~\cite{Zhu2021NICESLAM}, DF-Prior~\cite{Hu2023LNI-ADFP}, and Point-SLAM~\cite{Sandström2023ICCVpointslam}, due to higher quality renderings produced by view-tied Gaussians. Our view-tied Gaussians also show advantages over the original Gaussians used in SplaTAM~\cite{keetha2024splatam}, Gaussian-SLAM~\cite{yugay2023gaussianslam}, and GS-SLAM~\cite{yan2023gs} in terms of using much more Gaussians to represent local details in a more efficient manner, leading to better renderings to compare with the observations during tracking. However, relying on data-driven priors, LoopSplat~\cite{zhu2024_loopsplat} reported more accurate camera tracking in terms of average accuracy, while our method does not need any priors.

\begin{figure}[t]
  \centering
   \includegraphics[width=\linewidth]{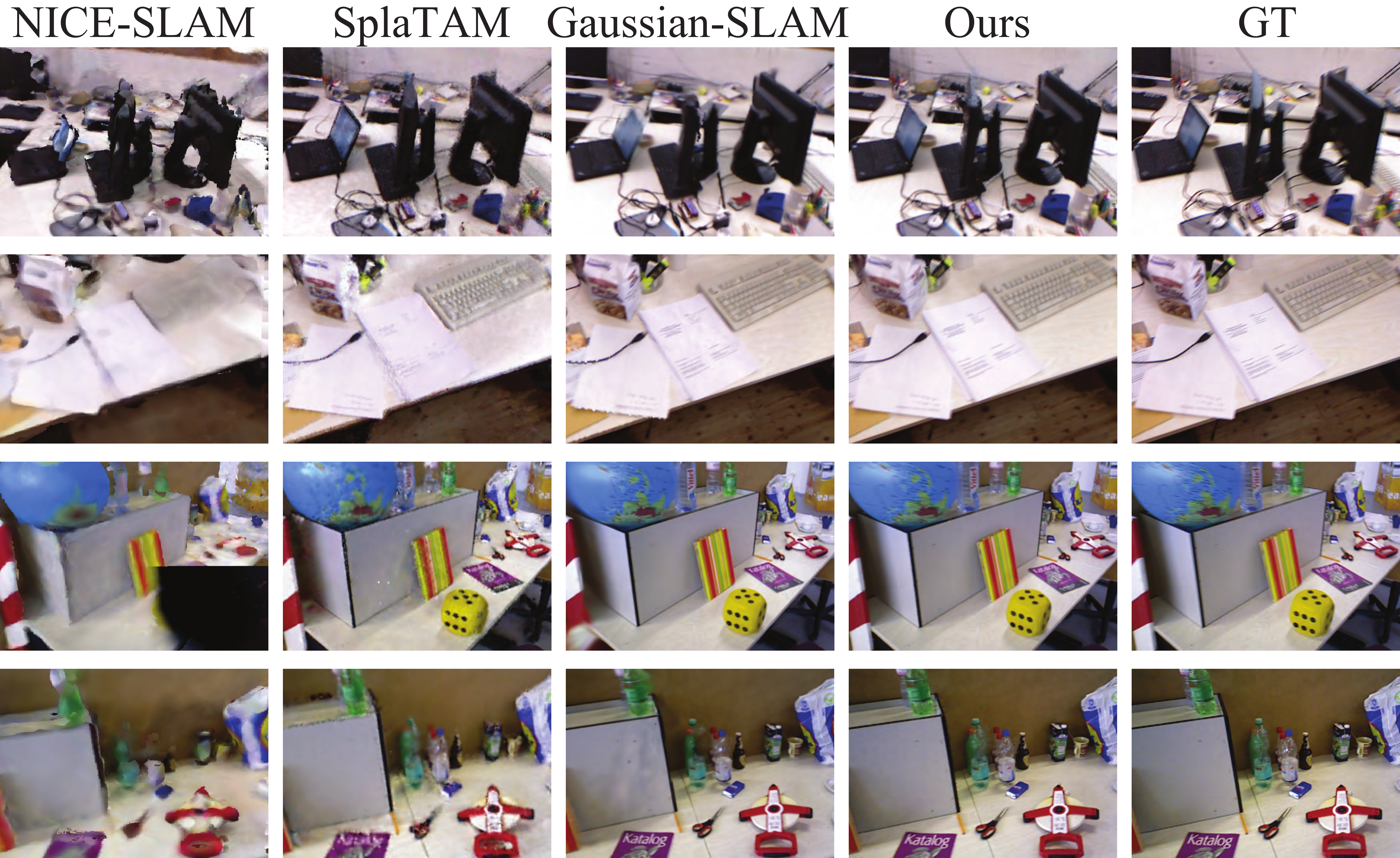}
   \vskip -0.1in
   \caption{Visual comparisons in rendering on TUM-RGBD.}
   \label{fig:Renderingtum}
\vskip -0.2in
\end{figure}

Due to the ability of using more Gaussians to describe local details, our method produces the best rendered images, as shown in Tab.~\ref{Tab:rendering}. Error map comparisons in Fig.~\ref{fig:TackingReplica} highlight our rendering quality, which is much better than others, especially in areas with sudden color changes. Because of better renderings, our methods also produce the most accurate reconstruction in Tab.~\ref{Tab:reconreplicaperscene}. We follow the previous method~\cite{Sandström2023ICCVpointslam} to render depth maps from the estimated camera poses, and fuse these depth maps into a TSDF for reconstruction. Visual comparisons in Fig.~\ref{fig:reconstructionReplica} show that we can recover more accurate geometry details.

\begin{figure}[!b]
  \centering
\vskip -0.2in
   \includegraphics[width=\linewidth]{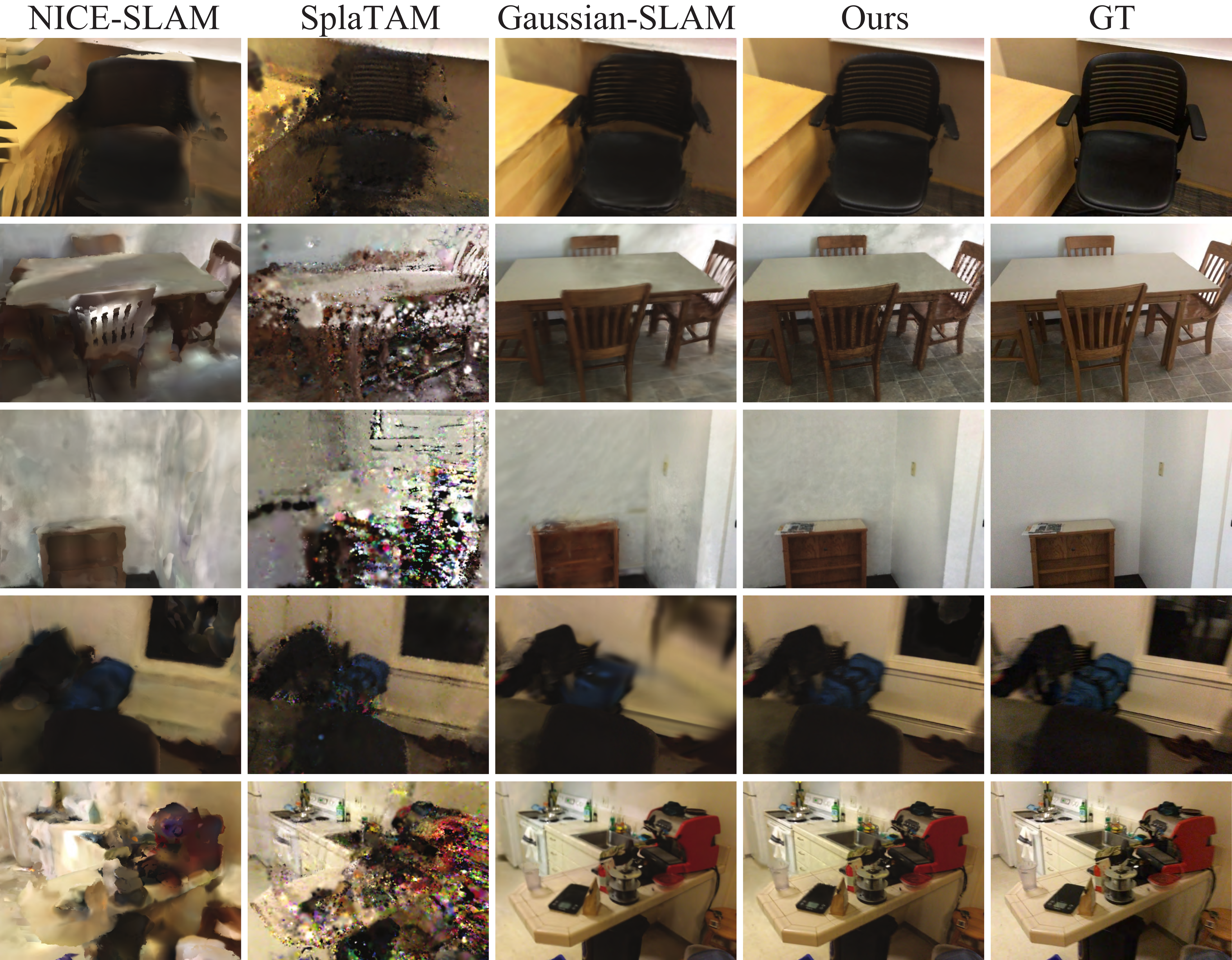}
   \vskip -0.1in
   \caption{Visual comparisons in rendering on ScanNet.}
   \label{fig:RenderingScanNet}
\end{figure}

\noindent\textbf{Results on TUM-RGBD. }We report our results on the TUM-RGBD dataset in camera tracking in Tab.~\ref{Tab:camtumperscene} and rendering in Tab.~\ref{Tab:rendering}. We follow previous methods~\cite{Zhu2021NICESLAM, keetha2024splatam,yugay2023gaussianslam, liso2024loopyslam,zhu2024_loopsplat,Sandström2023ICCVpointslam} and evaluate our method on the $3$ widely used scenes in TUM-RGBD.

\begin{figure*}[t!]
  \centering
   \includegraphics[width=\linewidth]{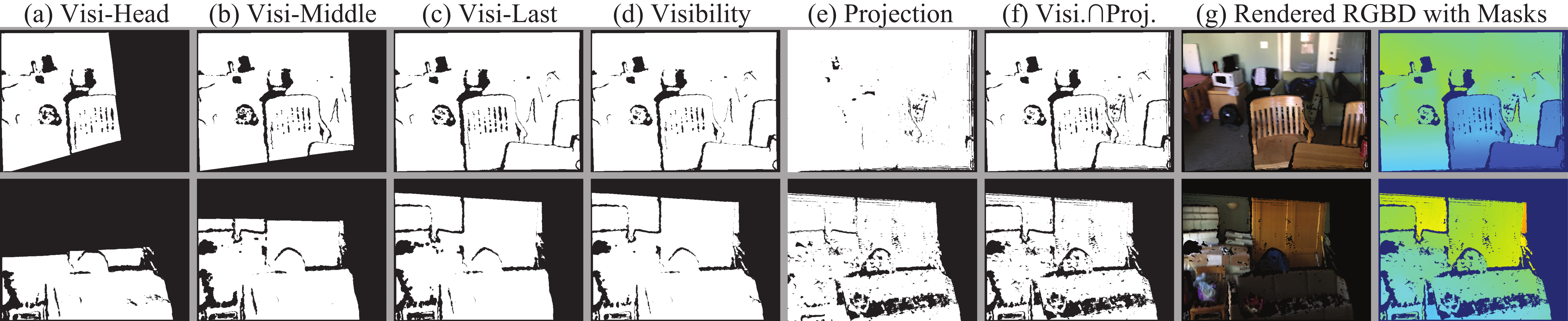}
   \vskip -0.1in
   \caption{Visualization of visibility and Gaussian projections on a head frame during tracking in a section.}
   \label{fig:visibility}
\end{figure*}

\begin{figure*}[!t]
  \centering
   \includegraphics[width=\linewidth]{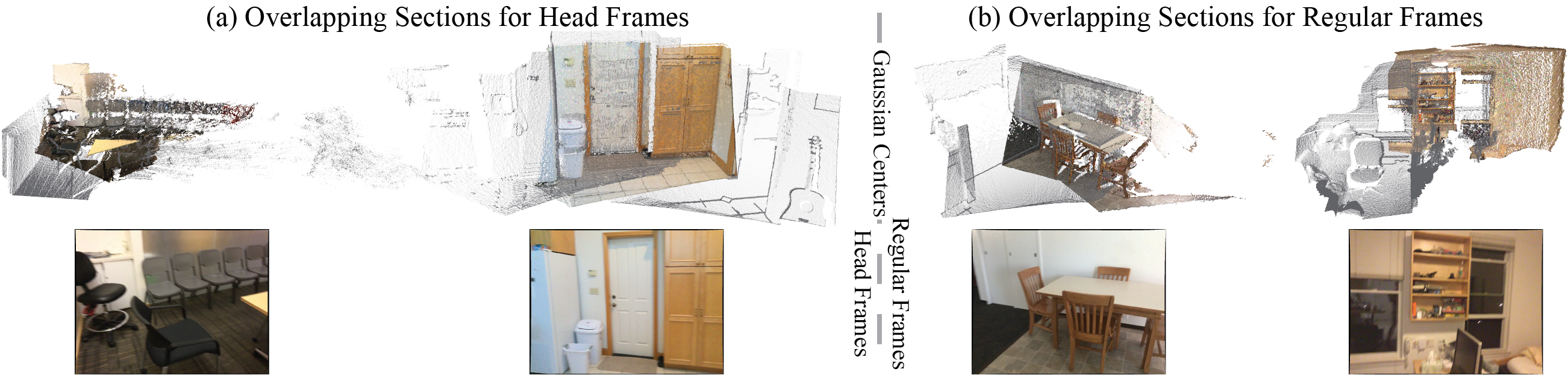}
   \vskip -0.05in
   \caption{Visualization of Gaussian centers with colors in the selected overlapping section for tracking (a) head frames and (b) regular frames. The Gaussian centers nearby are shown without color.}
   \label{fig:overlapssections}
\vskip -0.1in
\end{figure*}

\begin{table*}[!]
\caption{Tracking comparisons in ATE RMSE $\downarrow [\mathrm{cm}]$ on ScanNet. $*$ denotes use of pre-trained data-driven priors.}
\vskip 0.1in
  \centering
  \resizebox{\linewidth}{!}{
\begin{tabular}{ccccc!{\vrule width 1pt}cccccc}
\toprule
& \multicolumn{4}{c}{Neural Implicit Fields} & \multicolumn{5}{c}{3D Gaussian Splatting} \\
\midrule
Methods & NICE-SLAM & Vox-Fusion & Point-SLAM & Loopy-SLAM$*$ & SplaTAM &  Gaussian-SLAM & LoopSplat$*$ & CG-SLAM$*$ & Ours \\
\midrule
 Avg. & \textbf{10.7} & 26.9 & 12.2 & 7.7 & 11.9 & 15.4 & 7.7 & 8.1 & 11.3 \\

\bottomrule
\end{tabular}}
\label{Tab:camscannetperscene}
\vskip -0.1in
\end{table*}

\begin{table*}[!]
\caption{Tracking comparisons in ATE RMSE$\downarrow[\mathrm{cm}]$ on ScanNet++. $*$ denotes use of pre-trained data-driven priors.}
\vskip 0.1in
  \centering
  \resizebox{0.9\linewidth}{!}{
\begin{tabular}{cccc!{\vrule width 1pt}cccc}
\toprule
& \multicolumn{3}{c}{Neural Implicit Fields} & \multicolumn{4}{c}{3D Gaussian Splatting} \\
\midrule
Methods & Point-SLAM & ESLAM & Loopy-SLAM$*$ & SplaTAM &  Gaussian-SLAM & LoopSplat$*$ & Ours \\
\midrule
Avg. & 511.24 & 22.14 & 113.63 & 89.41 & 2.68 & 2.05 & \textbf{1.55}   \\

\bottomrule
\end{tabular}}
\label{Tab:camscannetppperscene}
\vskip -0.1in
\end{table*}

Comparisons in camera tracking in Tab.~\ref{Tab:camtumperscene} show that GS-based methods estimate camera poses more accurately than NeRF-based methods. Although the input RGBD observations are not in high resolution and with good quality, our method still produces the best tracking accuracy. Regarding the rendering, our view-tied Gaussians show even more advantages over the other methods in Tab.~\ref{Tab:rendering}. Visual comparisons in Fig.~\ref{fig:Renderingtum} highlight our improvement over the other methods. Compared to previous GS-based SLAM methods, our method can use many more Gaussians tied at each pixel on depth images to fit sudden color change without needing to maintain the consistency of Gaussians over the whole scene to a set of keyframes throughout the training, which enables Gaussians to focus more on local details.

\noindent\textbf{Results on ScanNet. }Our evaluations in camera tracking and mapping scenes with rendering views are reported in Tab.~\ref{Tab:camscannetperscene} and Tab.~\ref{Tab:rendering}, respectively. We produce the most accurate tracking performance in terms of average performance. Based on the camera poses, our method also significantly improves the rendering quality on ScanNet, as shown in Fig.~\ref{fig:RenderingScanNet}. The rendering improvement also justifies our advantages of using view-tied Gaussians on real-captured scenes. Besides good ability of recovering appearance details, with motion blur and low image quality in real images, our view-tied Gaussians can also limit these negative impact just on several neighboring views but not on all Gaussians in the scene during mapping. 

\noindent\textbf{Results on ScanNet++. }We report tracking results on the widely used $5$ scenes in ScanNet++ in Tab.~\ref{Tab:camscannetppperscene}. Compared to GS-based methods, our methods can estimate more accurate camera poses thanks to the more accurate renderings. Regarding novel view synthesis, please refer to our supplementary materials for more details.



\subsection{Ablation Studies and Analysis}
We justify the effectiveness of each design on synthetic and real scenes in Replica~\cite{replica} and TUM-RGBD~\cite{6385773ATERMSE_tumrgbd}. 

\noindent\textbf{3D Gaussians. }We conduct experiments to highlight the effect of view-tied Gaussians in Tab.~\ref{Tab:ablation3dgs}. We replace our Gaussians with ellipsoid Gaussians with either fixed (``aniso + w/ VT'') or learnable locations (``aniso + w/o VT''). We also show the effect of learnable locations with our simplified Gaussians (``iso + w/o VT''). Without tying Gaussians to depth maps, we need to store Gaussian locations, which limits the number of Gaussians we can use, degenerating the rendering quality. 
The comparisons show that our view-tied Gaussians not only significantly reduce the size of each Gaussian (number of parameters) but also achieve good rendering quality with our tracking and mapping strategies.

\begin{table*}[t]
\vskip -0.1in
\caption{Ablation study on attributes of 3D Gaussians (aniso: anisotropic Gaussians, iso: isotropic Gaussians, VT: view-tied Gaussians).}
\vskip 0.1in
  \centering
  \resizebox{0.7\linewidth}{!}{
\begin{tabular}{lcccc}
\toprule
Metric & aniso + w/o VT & aniso+ w/ VT & iso + w/o VT & iso + w/ VT (Ours) \\
\midrule
PSNR$\uparrow [\mathrm{dB}]$ & 37.62 & 38.46 & 36.73 & \textbf{39.95} \\
ATE RMSE $\downarrow [\mathrm{cm}]$ & 19.60 & 25.65 & 21.45 & \textbf{0.22} \\
\midrule
Param./Gaussian & 14 & 11 & 8 & \textbf{5} \\

\bottomrule
\end{tabular}}
\label{Tab:ablation3dgs}
\vskip -0.1in
\end{table*}

\begin{table*}[t]
\caption{Ablation study on the length of section $\bm{S}$, overlap selecting strategy, and visible mask. 
} 
\vskip 0.1in
  \centering
  \resizebox{\linewidth}{!}{
\begin{tabular}{lccccc!{\vrule width 1pt}cccc!{\vrule width 1pt}cc}
\toprule
 & \multicolumn{5}{c}{Length of section $\bm{S}$} & \multicolumn{4}{c}{Overlap section selecting strategy} &\multicolumn{2}{c}{Visibility mask} \\
\midrule
Metric & 20 & 40 (Ours) & 60 & 80 & 100 & nearest & largest & multiple & Ours & w/o & w/  (Ours) \\
\midrule
PSNR$\uparrow [\mathrm{dB}]$ & 39.92 & \textbf{39.95} & 39.34 & 38.87 & 38.56 & 38.52 & 38.74 & 38.03 & \textbf{39.95} & 30.40 & \textbf{30.51} \\
ATE RMSE $\downarrow [\mathrm{cm}]$ & 0.25 & \textbf{0.22} & 0.23 & 0.26 & 0.24 & 0.30 & 0.34 & 0.28 & \textbf{0.22} & 6.8 & \textbf{4.4} \\

\bottomrule
\end{tabular}}
\label{Tab:ablationsectionlengthandoverlapandvismask}
\vskip -0.1in
\end{table*}

\begin{table*}[!]
\caption{Runtime and Memory Usage on Replica.} 
\vskip 0.1in
  \centering
  \resizebox{0.8\linewidth}{!}{
\begin{tabular}{lccccc}
\toprule
 Method & NICE-SLAM & Point-SLAM & SplaTAM & Gaussian-SLAM & Ours \\
\midrule
Tracking/Frame(s) & 1.06 & 1.11 & 2.70 & 0.83 & 1.92 \\
Mapping/Frame(s) & 1.15 & 3.52 & 4.89 & 0.93 & 2.43 \\
\midrule
Total Num of Gaussians & - & - & 5832$K$ & 32592$K$ & 97823$K$ \\
Max Num of Gaussians & - & - & 5832$K$ & 1983$K$ & 2664$K$ \\
\bottomrule
\end{tabular}}
\label{Tab:memoryandtime}
\vskip -0.1in
\end{table*}

\noindent\textbf{Section Length. }The number of frames in a section is also a factor impacting the performance. Tab.~\ref{Tab:ablationsectionlengthandoverlapandvismask} shows that too few or too many frames in one section will degenerate the performance if we do not adjust other parameters like optimization iterations during tracking or mapping. Too few frames will increase the possibility of cumulating camera pose errors while changing into the next section. Instead, too many frames will need more iterations during mapping to learn Gaussians well. 
We cannot use a large number of Gaussians if setting the length of the section as 1, which degenerates the rendering and tracking performance. We visualize some overlapping sections during tracking in Fig.~\ref{fig:overlapssections}.

\noindent\textbf{Overlapping Section Selection Strategy. }Our strategy of selecting overlapping sections is also important for good renderings in tracking the head frame. Our selection based on visibility and preference to the most front sections significantly reduces the impact of pose error cumulation. We compare this strategy with selecting the nearest section (``nearest''), the section with the largest overlaps (``largest''), or multiple sections (``multiple'') that add the nearest section to the selection we selected. The comparisons in Tab.~\ref{Tab:ablationsectionlengthandoverlapandvismask} show that our selection strategy achieves the best performance.

\noindent\textbf{Visibility. }We consider visibility in both tracking and mapping to reduce the error brought by unseen or occluded areas. Using no visibility in the loss function will degenerate the performance, as shown in Tab.~\ref{Tab:ablationsectionlengthandoverlapandvismask}, since the error in the unseen area will not be minimized by adjusting camera poses or optimizing Gaussian attributes. We visualize the visibility masks and projection masks in Fig.~\ref{fig:visibility}. The visibility masks of a head frame to the first, the middle, and the last frame in the selected overlapping section are shown in Fig.~\ref{fig:visibility} (a)-(c), respectively. The overall visibility to the overlapping section is the union of these 3 visibility masks in Fig.~\ref{fig:visibility} (d). The silhouette produced by projections of Gaussians in the same section is shown in Fig.~\ref{fig:visibility} (e). We use the intersection of visibility mask and silhouette mask in Fig.~\ref{fig:visibility} (f) to weight rendered RGBD images in Fig.~\ref{fig:visibility} (g).

\noindent\textbf{Runtime, Storage, and Scalability. }We report the average time for tracking and mapping one frame in Tab.~\ref{Tab:memoryandtime} on a single NVIDIA RTX4090. Our runtime is comparable to other GS-based methods. Regarding the number of Gaussians, we show a great advantage over other methods. We can initiate many more Gaussians over the whole scene than other GS-based methods for better rendering. Meanwhile, around each frame, we still have good control of the number of Gaussians so that we can maximize the usage of the limited GPU memory, which finds a good balance between scalability and limited hardware resources. This enables us to handle much larger scenes in more frames with more Gaussians than other SLAM methods.

\section{Conclusion}
We propose VTGaussian-SLAM to improve the performance of SLAM with 3D Gaussian splatting in terms of rendering quality, tracking accuracy, and scalability. We showed that our view-tied Gaussians can significantly save storage so that we can maintain a large amount of these Gaussians in the limited GPU memory for either higher rendering quality or larger areas. Our tracking and mapping strategies take good advantage of these benefits, which allows us to merely maintain and optimize Gaussians that contribute to the most recent views a lot. This ensures the high rendering quality for the latest views, and leads to more accurate camera tracking and mapping. We justified the effectiveness of each design and reported visual and numerical evaluations to illustrate our advantages over the latest SLAM methods.

\section*{Acknowledgements}
This project was partially supported by an NVIDIA academic award and a Richard Barber research award.

\section*{Impact Statement}
This paper presents work whose goal is to advance the field of 
Machine Learning. There are many potential societal consequences 
of our work, none which we feel must be specifically highlighted here.


\bibliography{papers}
\bibliographystyle{icml2025}

\newpage
\clearpage
\appendix

\renewcommand{\thesubsection}{\Alph{subsection}}
\section*{Supplementary Material}

In this supplementary material, we will cover more details about the implementation and results on each scene. Additionally, we also show more results regarding Novel View Synthesis on ScanNet++~\cite{yeshwanthliu2023scannetpp}.

\subsection{Implementation Details}
We implemented VTGaussian-SLAM in Python using the PyTorch framework, and ran all experiments on NVIDIA RTX4090 GPUs. During mapping, we will initialize view-tied Gaussians from input RGBD images. Following ~\cite{keetha2024splatam}, we initialized the radius of each Gaussian $r$ using the following equation:
\begin{equation}
    r = \frac{\bm{D_{GT}}}{f},
\end{equation}
Here $\bm{D_{GT}}$ is the ground-truth depth, and $f$ is the focal length. As for the learning rate of 3D Gaussians, we set $lr_{color} = 0.0025$ for the color, $lr_{radius}=0.005$ for the radius, and $lr_{opacity}=0.05$ for the opacity separately. For camera tracking, we initialize the pose using the constant speed assumption on Replica~\cite{replica}, TUM-RGBD~\cite{6385773ATERMSE_tumrgbd}, and ScanNet~\cite{scannet}. Specifically, on ScanNet++~\cite{yeshwanthliu2023scannetpp}, since it is not a dataset originally designed for SLAM tasks, some sudden large motion changes are occurring in the DSLR-captured sequences, we follow previous methods~\cite{zhu2024_loopsplat, yugay2023gaussianslam} to utilizing multi-scale RGBD odometry~\cite{colorptreg_odo} to help the pose initialization if the rendering error with the pose initialized by constant speed assumption is 50 times larger than the average of the rendering loss for previous frames after the tracking optimization. We set the learning rate of pose to $lr_{rot} = 0.0004$ and $lr_{trans}=0.002$ on Replica, $lr_{rot} = 0.002$ and $lr_{trans}=0.002$ on TUM-RGBD and ScanNet, and $lr_{rot} = 0.001$ and $lr_{trans}=0.01$ on ScanNet++ separately. Additionally, we set the overlapping threshold $\gamma = 0.26$ on TUM-RGBD, and $\gamma = 0.24$ on ScanNet and ScanNet++ separately.

\subsection{More Results}
\noindent\textbf{Per-scene Results. }We present more detailed results on each scene in Replica~\cite{replica}, TUM-RGBD~\cite{6385773ATERMSE_tumrgbd}, ScanNet~\cite{scannet}, and ScanNet++~\cite{yeshwanthliu2023scannetpp}. Replica is a synthetic dataset, whereas TUM-RGBD, ScanNet, and ScanNet++ are real-world datasets. TUM-RGBD were captured using an external motion capture system, while ScanNet uses poses from BundleFusion~\cite{dai2017bundlefusion}, and ScanNet++ utilizes a laser scan to register the images for acquiring corresponding camera poses. We follow previous methods~\cite{keetha2024splatam,yugay2023gaussianslam,MatsukiCVPR2024_monogs,zhu2024_loopsplat,wei2024gsfusiononlinergbdmapping,li2024sgsslamsemanticgaussiansplatting} and conduct experiments on three scenes of TUM-RGBD, six scenes of ScanNet, and five scenes of ScanNet++ ((a) \texttt{b20a261fdf}, (b) \texttt{8b5caf3398}, (c) \texttt{fb05e13ad1}, (d) \texttt{2e74812d00}, (e) \texttt{281bc17764}) to evaluate our performance.

We report numerical comparisons in camera tracking in each scene in Replica~\cite{replica} in Tab.~\ref{Tab:suppcamreplicaperscene}, in TUM-RGBD~\cite{6385773ATERMSE_tumrgbd} in Tab.~\ref{Tab:suppcamtumperscene}, in ScanNet~\cite{scannet} in Tab.~\ref{Tab:suppcamscannetperscene}, and in ScanNet++~\cite{yeshwanthliu2023scannetpp} in Tab.~\ref{Tab:suppcamscannetppperscene}. The comparisons show that our methods can estimate more accurate camera poses in most scenes.

\begin{figure*}
  \centering
   \includegraphics[width=\linewidth]{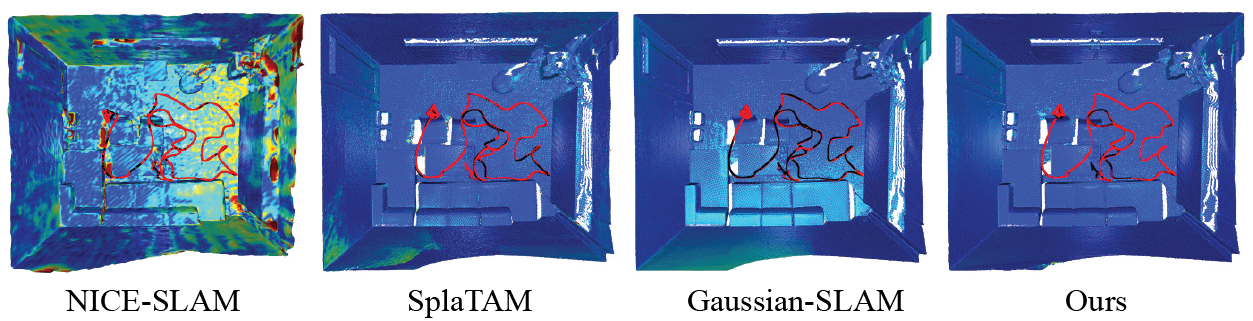}
   \caption{Visual comparisons in camera tracking on Replica. We also show error maps on reconstructions. Please refer to our video for a more complete comparison during scanning.}
   \label{fig:suppvistracking}
\end{figure*}

Moreover, we report reconstruction comparisons in each scene in Replica~\cite{replica} in Tab.~\ref{Tab:suppreconreplicaperscene}. We utilize depth L1 and F1-score as metrics to evaluate the mesh obtained by marching cubes~\cite{Lorensen87marchingcubes} following a similar procedure in \cite{Sandström2023ICCVpointslam}. Compared to the latest methods, our methods can recover more accurate geometry, although Point-SLAM~\cite{Sandström2023ICCVpointslam} requires ground truth depth
images as input to guide sampling when rendering, which is an unfair advantage compared to other methods. Specifically, we also show visual comparisons regarding camera tracking and reconstruction in Fig.~\ref{fig:suppvistracking}. Please refer to our supplementary video for more details of this comparison.

We also report comparisons in rendering in each scene from the training views in Replica~\cite{replica}, TUM-RGBD~\cite{6385773ATERMSE_tumrgbd}, ScanNet~\cite{scannet}, and ScanNet++~\cite{yeshwanthliu2023scannetpp} separately. Due to our view-tied strategy, there will be more 3D Gaussians used for rendering, which leads to better rendering quality compared to the latest methods as shown in Tab.~\ref{Tab:supprenderreplicaperscene}, Tab.~\ref{Tab:supprendertumperscene}, Tab.~\ref{Tab:supprenderscannetperscene}, and Tab.~\ref{Tab:supprenderscannetppperscene}.

\noindent\textbf{Large-scale scenes Results. }We additionally report our performance on extremely large scenes, such as city-level scenes in KITTI~\cite{Geiger2012CVPR}. Since many moving objects exist in KITTI sequences, we only select part of the sequences to evaluate our tracking and rendering performance in Tab.~\ref{Tab:suppkittitracking} and Tab.~\ref{Tab:suppkittimapping}. The comparisons show that our methods can estimate more accurate camera poses with high quality rendering performance. Meanwhile, we report memory consumption on KITTI in Tab.~\ref{Tab:suppkittimapping}. Each method uses the most Gaussians until no improvement can be made. We use a little bit more memory, but we manage to use more Gaussians to produce much better rendering.

\subsection{Novel View Synthesis}
\begin{figure*}
  \centering
   \includegraphics[width=\linewidth]{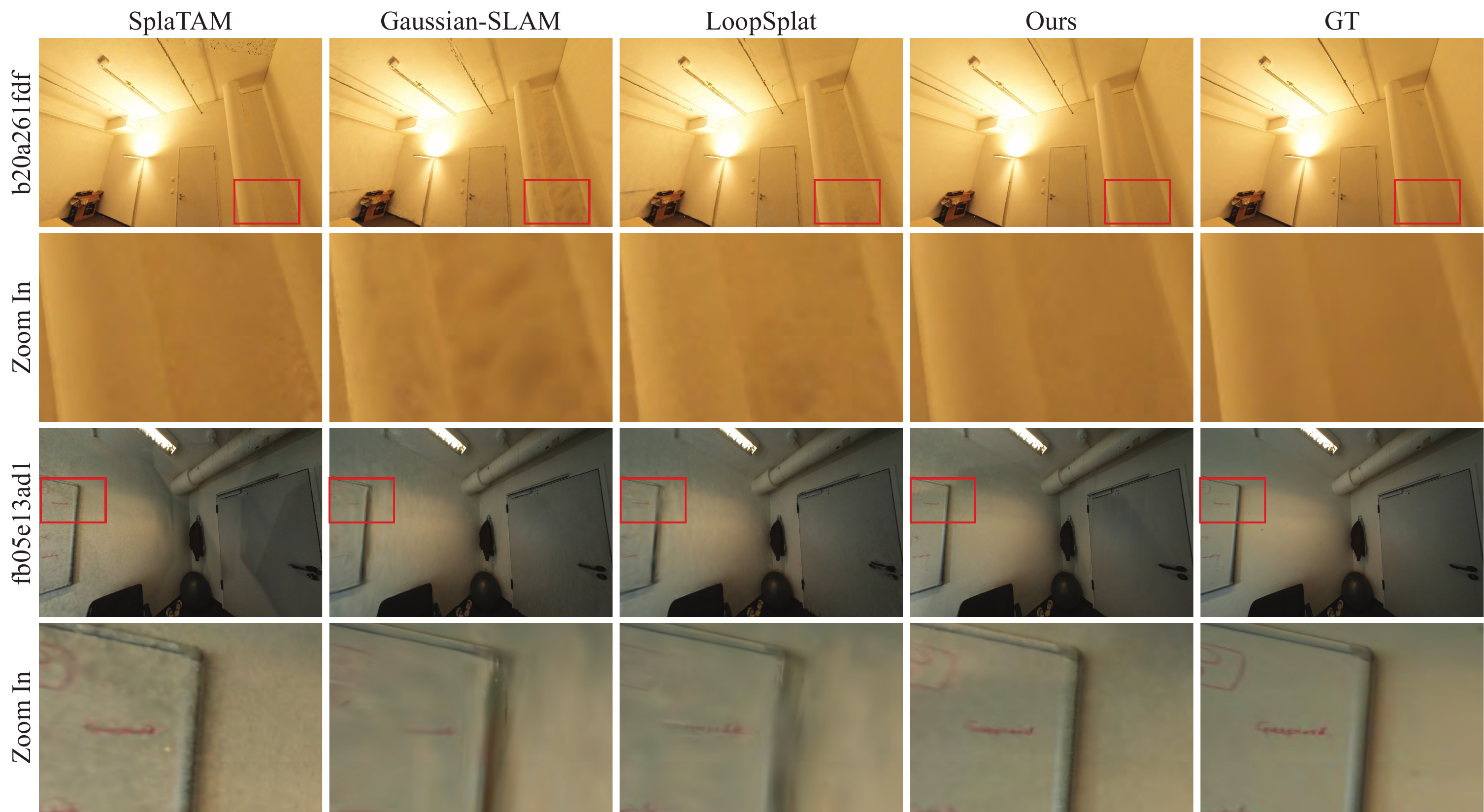}
   \caption{Visual comparisons in training view rendering on ScanNet++~\cite{yeshwanthliu2023scannetpp}.}
   \label{fig:supprenderscannetpp}
\end{figure*}

\begin{figure*}
  \centering
   \includegraphics[width=\linewidth]{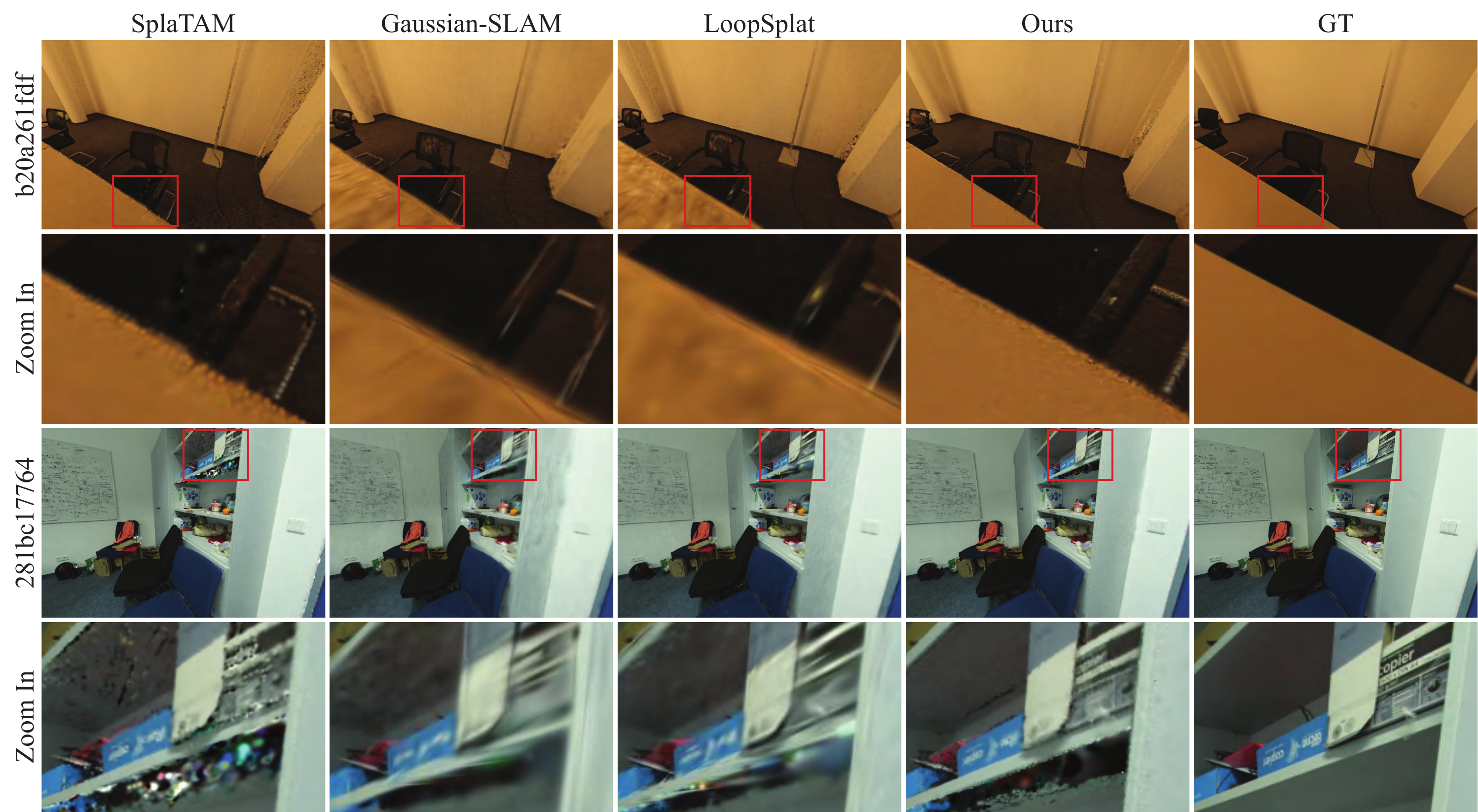}
   \caption{Visual comparisons in novel view rendering on ScanNet++~\cite{yeshwanthliu2023scannetpp}.}
   \label{fig:suppnvsrenderscannetpp}
\end{figure*}

We evaluate the performance in novel view synthesis on ScanNet++~\cite{yeshwanthliu2023scannetpp}. The testing views in ScanNet++ are from held-out views, which are much different from the training views. To evaluate PSNR on all test views, we use a post-processing step after training is finished. In post-processing, we use all training views to refine all trained sections $\{{\bm{S}_k\}}$. Specifically, for each training view, we find all its overlapping sections 
and refine all these sections using this view. Compared to the latest methods \cite{zhu2024_loopsplat,yugay2023gaussianslam}, which take $10K$ iterations to refine their global map, our refinement only needs $1K$ iterations for novel view synthesis. In testing, given a novel view, we will find all overlapping sections $\bm{S}_k$, and concatenate all these sections to render the novel view image. Tab.~\ref{Tab:suppnvsrenderscannetppperscene} shows that our methods can obtain comparable novel view synthesis results to the latest methods. We also 
show qualitative rendering results of the training views in Fig.~\ref{fig:supprenderscannetpp} and novel view synthesis in Fig.~\ref{fig:suppnvsrenderscannetpp}.

\subsection{More analysis and visualization}
\subsubsection{Noise depth}
\begin{figure*}
  \centering
   \includegraphics[width=\linewidth]{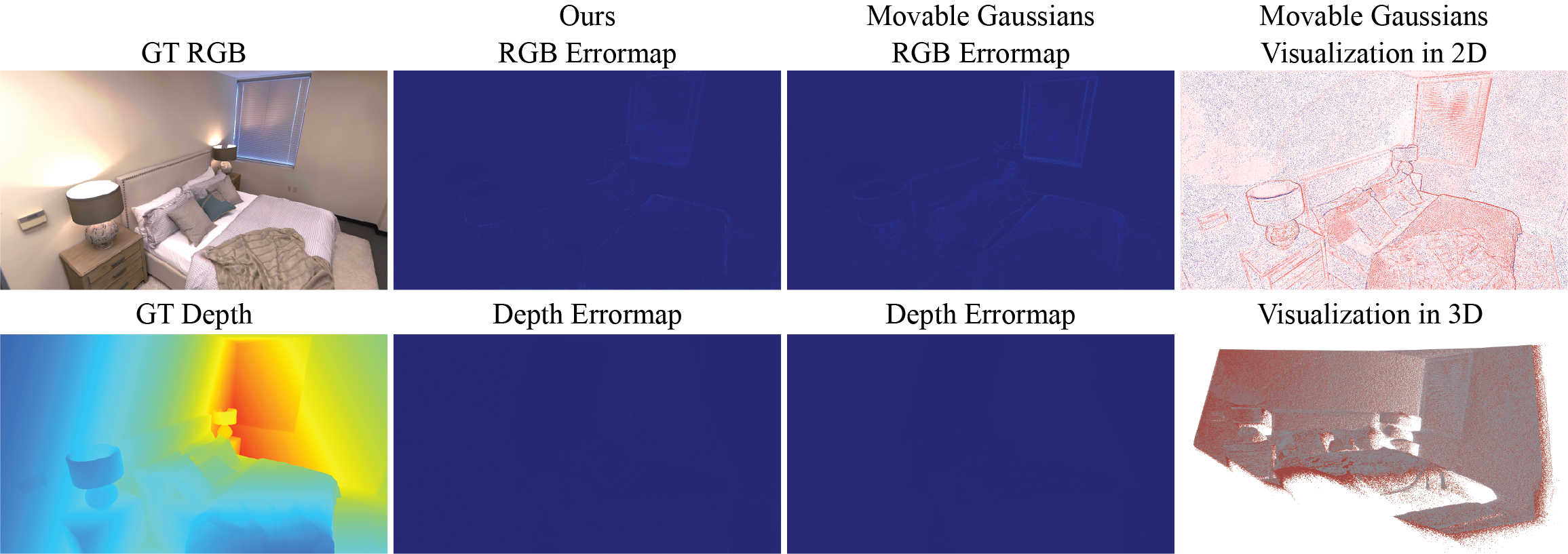}
   \caption{Visual comparisons on fixed Gaussians or movable Gaussians
(along the ray) in rendering on Replica. We provide rendering results, including RGB error maps and depth error maps rendered 
by using either our view-tied Gaussians or movable Gaussians along the ray. Additionally, we provide a 2D visualization of the optimized 
movable Gaussians (blue: movement backward along the ray, red: movement forward along the ray. Furthermore, we provide a 3D visualization
of the optimized movable Gaussians (gray: central points of Gaussians fixed at depth, red: central points of Gaussians optimized along the ray).}
   \label{fig:suppmovegs}
\end{figure*}

Our view-tied Gaussians can also resist the impact brought by noise in depth. 
Although Gaussians are fixed at depth with noise, Gaussian splatting is flexible enough to overfit the current frame and neighboring frames by tuning other attributes like color, opacity, and shape. Our results show that depth noises do not significantly impact the rendering performance. Meanwhile, we try to optimize the position of Gaussians along the ray direction, but we do not find an obvious improvement in rendering performance. We report additional results in Tab.~\ref{Tab:suppdepthnoise}. We also report a visual comparison using either fixed Gaussians or movable Gaussians (along the ray) in Fig.~\ref{fig:suppmovegs}.

\subsubsection{Issue of pose error cumulation}
\begin{figure}
  \centering
   \includegraphics[width=\linewidth]{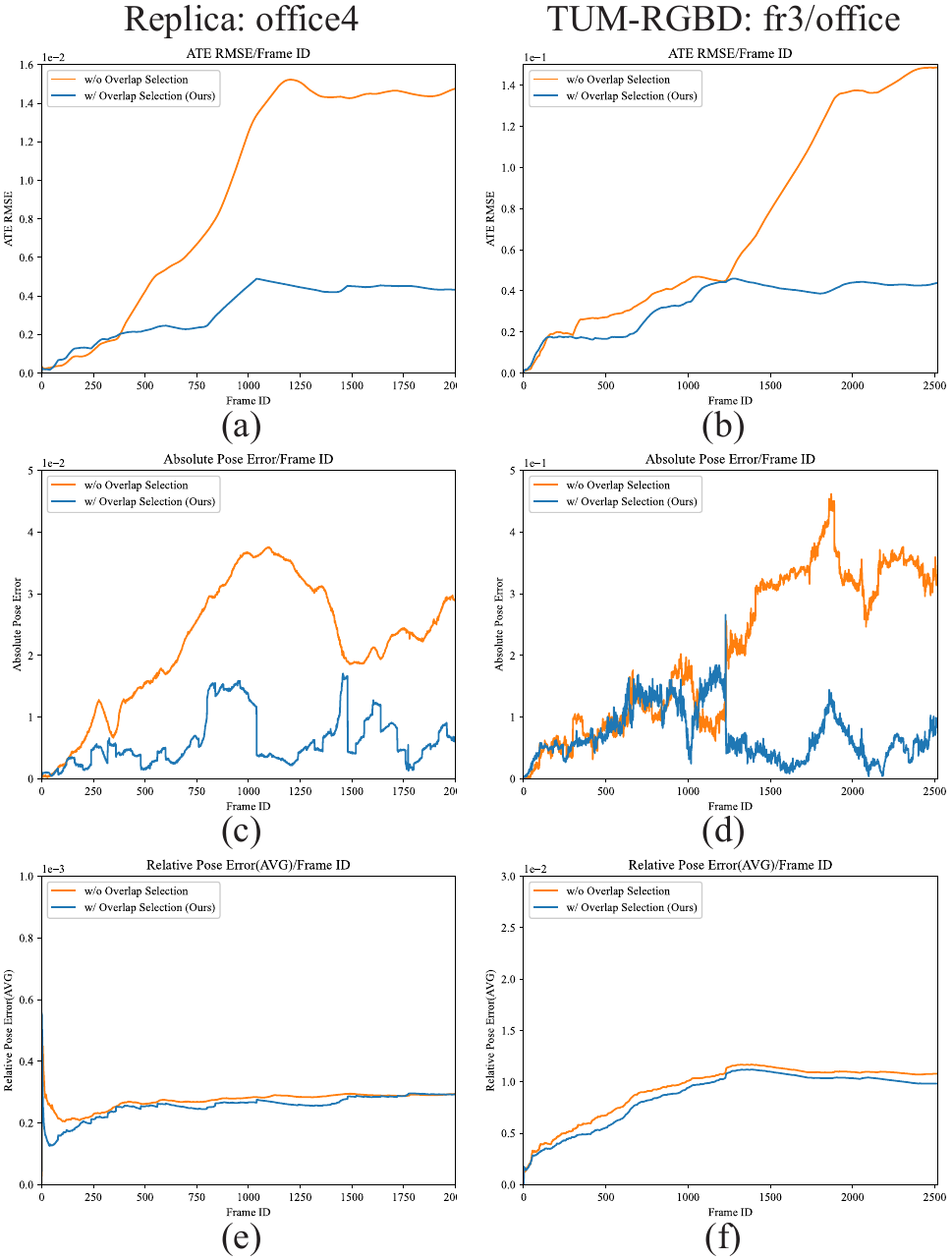}
   \caption{Comparison on w/ or w/o overlap selection when tracking.}
   \label{fig:suppabscurve_supp}
\end{figure}

Here we present the effectiveness of our tracking strategy. As shown by average pose accuracy (ATE RMSE) in Fig.~\ref{fig:suppabscurve_supp} (a) and (b), using overlapping selection for rendering in tracking can prevent the estimated camera pose from drifting away from the trajectory caused by the error cumulation. The absolute pose error at each frame is shown in Fig.~\ref{fig:suppabscurve_supp} (c) and (d). Since our method can render pretty good images, we produce small pose errors relative to the previous view in Fig.~\ref{fig:suppabscurve_supp} (e) and (f).

\subsubsection{Visualization of 3D Gaussians}
\begin{figure}
  \centering
   \includegraphics[width=\linewidth]{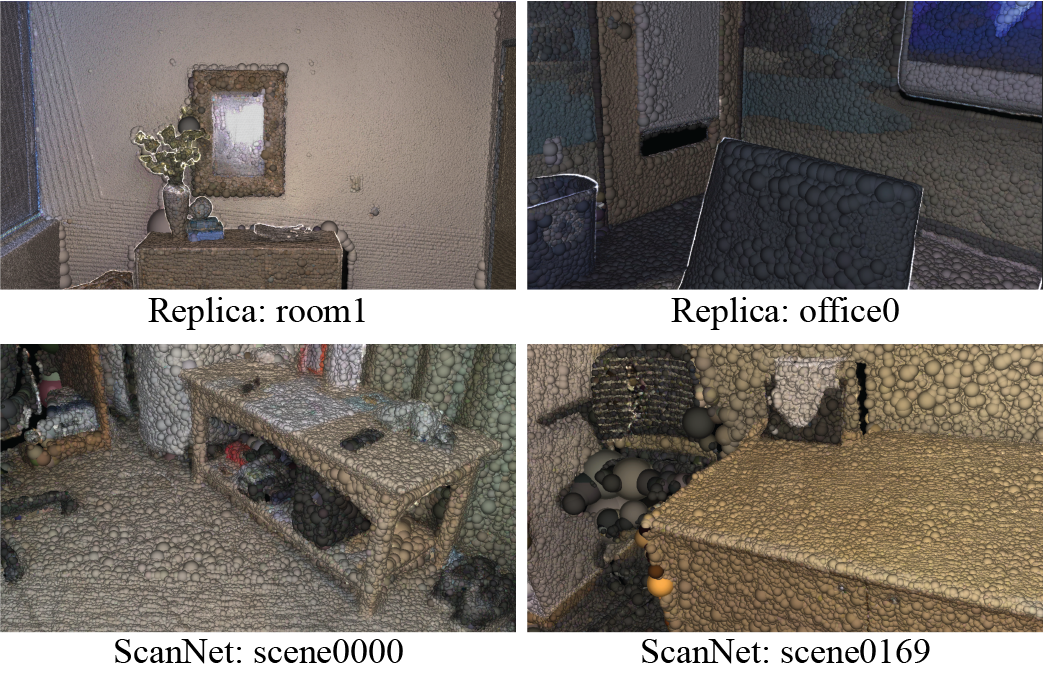}
   \caption{Visualization of optimized 3D Gaussians.}
   \label{fig:supp3dgsvis}
\end{figure}

Fig.~\ref{fig:supp3dgsvis} visualizes optimized 3D Gaussians. Our methods can initialize much denser 3D Gaussians to represent the whole scene due to the view-tied strategy. With more 3D Gaussians, our methods can render more realistic images. 

\begin{figure}
  \centering
   \includegraphics[width=\linewidth]{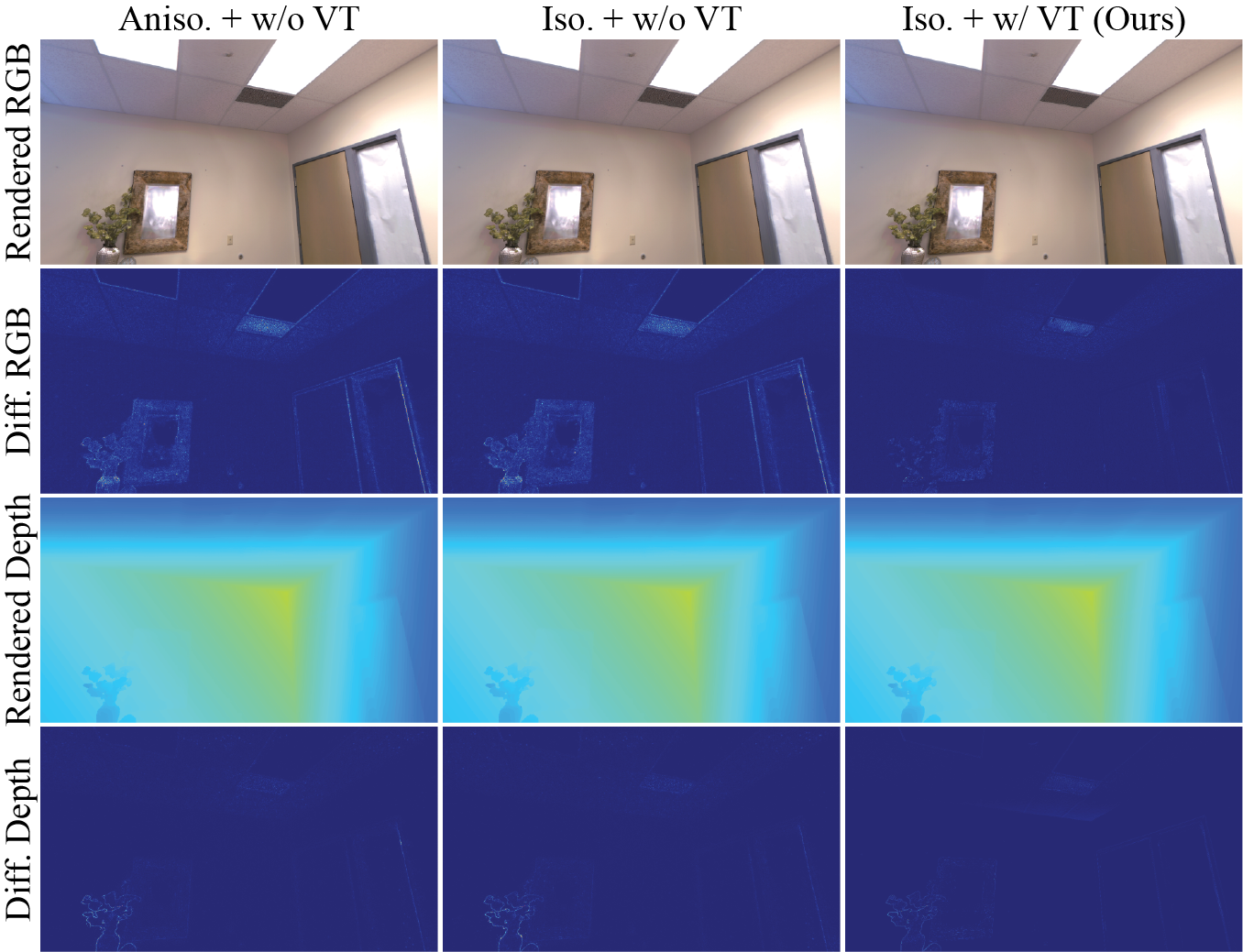}
   \caption{Comparisons of different kinds of Gaussians (with the same number). Our view-tied Gaussians in the 3rd column can recover more accurate RGB color and depth in renderings, while the ellipsoid Gaussians in the 1st column and sphere Gaussians in the 2nd column produce worse rendering quality with adjustable Gaussians. Please refer to our video for a complete comparison of optimization.}
   \label{fig:suppGaussianComp}
\end{figure}

We also report a comparison of different kinds of Gaussians in Fig.~\ref{fig:suppGaussianComp}. We employ the same number of Gaussians, but highlight the performance of our Gaussians in rendering with different specifications, such as using ellipsoid Gaussian (``Aniso.'') and learnable Gaussian locations (``w/o VT''). The comparison shows that our Gaussians can produce the minimum rendering errors.

\begin{figure}
  \centering
   \includegraphics[width=\linewidth]{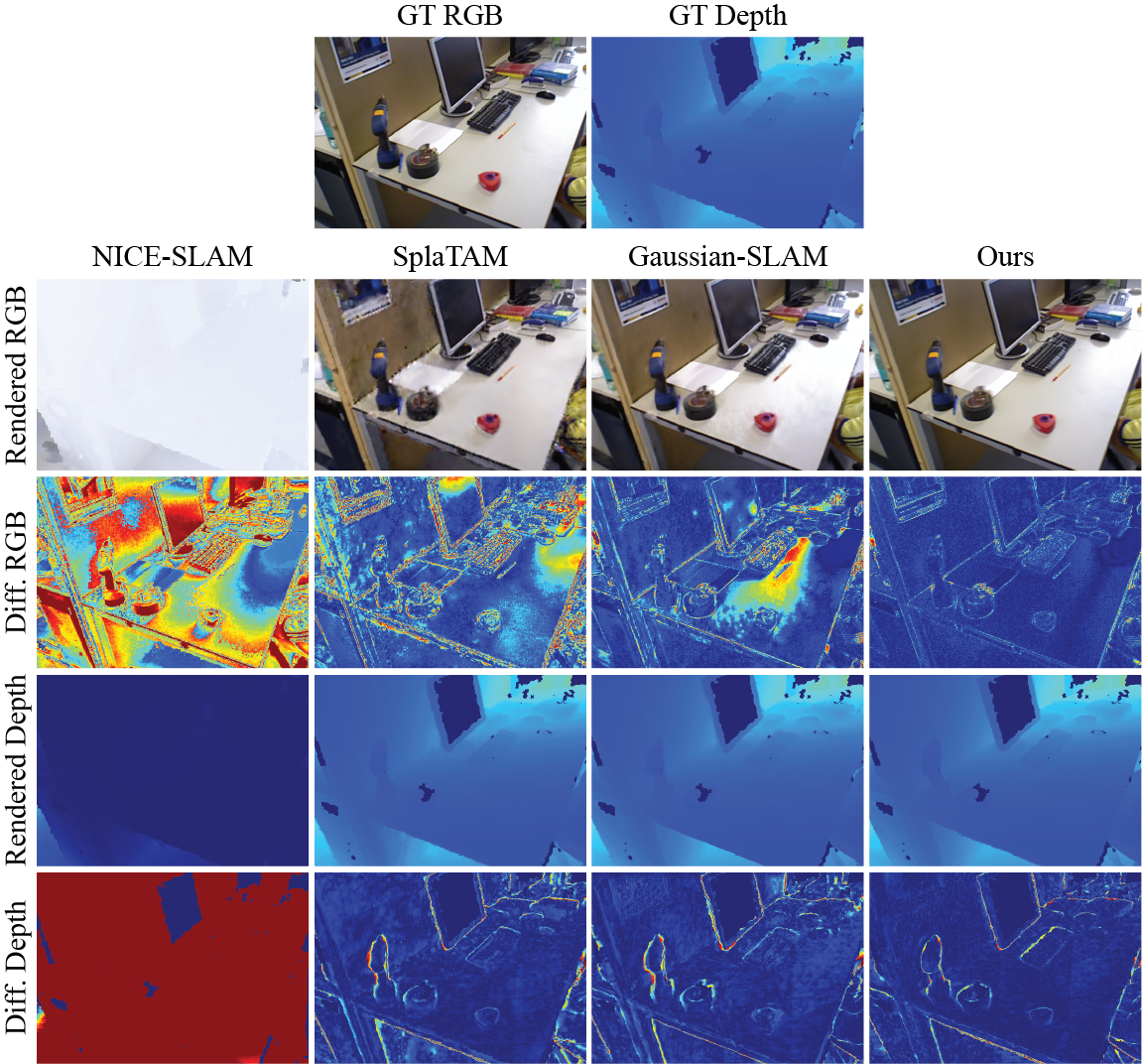}
   \caption{Visual comparisons of rendered images and depths. We also show error maps (large rendering errors are shown in red). Please refer to our video for more visual comparisons of rendered images.}
   \label{fig:supptrainingviewtum}
\end{figure}

Many more advantages can be shown with our estimated camera poses during mapping in Fig.~\ref{fig:supptrainingviewtum}. Our method can recover more details of the scene and produce more accurate renderings.

\subsection{Code}
Please see our project page for code at \href{https://machineperceptionlab.github.io/VTGaussian-SLAM-Project}{https://machineperceptionlab.github.io/VTGaussian-SLAM-Project}.

\subsection{Video}
We present more visualization in our video, such as visual comparisons and visualization of the optimization. Please watch our video for more details.

\begin{table*}
\caption{Tracking performance comparisons in ATE RMSE $\downarrow [\mathrm{cm}]$ on Replica~\cite{replica}. $*$ indicates methods relying on pre-trained data-driven priors.}
\vskip 0.1in
  \centering
  \resizebox{\linewidth}{!}{
\begin{tabular}{lccccccccc}
\toprule
Method & \texttt{Rm0} & \texttt{Rm1} & \texttt{Rm2} & \texttt{Off0} & \texttt{Off1} & \texttt{Off2} & \texttt{Off3} & \texttt{Off4} & Avg. \\
\midrule
\rowcolor{gray!20}
\multicolumn{10}{l}{\textit{Neural Implicit Fields}} \\
NICE-SLAM~\cite{Zhu2021NICESLAM} & 1.69 & 2.04 & 1.55 & 0.99 & 0.90 & 1.39 & 3.97 & 3.08 & 1.95 \\
DF-Prior~\cite{Hu2023LNI-ADFP} & 1.39 & 1.55 & 2.60 & 1.09 & 1.23 & 1.61 & 3.61 & 1.42 & 1.81 \\
Vox-Fusion~\cite{Yang_Li_Zhai_Ming_Liu_Zhang_2022_voxfusion} & \textbf{0.27} & 1.33 & 0.47 & 0.70 & 1.11 & \textbf{0.46} & \textbf{0.26} & \textbf{0.58} & 0.65 \\
ESLAM~\cite{johari-et-al-2023-ESLAM} & 0.71 & 0.70 & 0.52 & 0.57 & 0.55 & 0.58 & 0.72 & 0.63 & 0.63 \\
Point-SLAM~\cite{Sandström2023ICCVpointslam} & 0.61 & \textbf{0.41} & \textbf{0.37} & \textbf{0.38} & \textbf{0.48} & 0.54 & 0.72 & 0.63 & \textbf{0.52} \\
\hdashline
Loopy-SLAM$*$~\cite{liso2024loopyslam} & 0.24 & 0.24 & 0.28 & 0.26 & 0.40 & 0.29 & 0.22 & 0.35 & 0.29 \\
\midrule
\midrule
\rowcolor{gray!20}
\multicolumn{10}{l}{\textit{3D Gaussian Splatting}} \\
SplaTAM~\cite{keetha2024splatam} & 0.31 & 0.40 & 0.29 & 0.47 & 0.27 & \textbf{0.29} & 0.32 & 0.55 & 0.36 \\
GS-SLAM~\cite{yan2023gs} & 0.48 & 0.53 & 0.33 & 0.52 & 0.41 & 0.59 & 0.46 & 0.70 & 0.50 \\
Gaussian-SLAM~\cite{yugay2023gaussianslam} & 0.29 & 0.29 & 0.22 & 0.37 & \textbf{0.23} & 0.41 & 0.30 & \textbf{0.35} & 0.31 \\
\hdashline
LoopSplat$*$~\cite{zhu2024_loopsplat} & 0.28 & 0.22 & 0.17 & 0.22 & 0.16 & 0.49 & 0.20 & 0.30 & 0.26 \\
CG-SLAM$*$~\cite{hu2024cg} & 0.29 & 0.27 & 0.25 & 0.33 & 0.14 & 0.28 & 0.31 & 0.29 & 0.27 \\
\midrule
\midrule
Ours & \textbf{0.22} & \textbf{0.26} & \textbf{0.19} & \textbf{0.28} & 0.26 & 0.34 & \textbf{0.25} & 0.43 & \textbf{0.28} \\

\bottomrule
\end{tabular}}
\label{Tab:suppcamreplicaperscene}
\end{table*}

\begin{table*}
\caption{Tracking performance comparisons in ATE RMSE $\downarrow [\mathrm{cm}]$ on TUM-RGBD~\cite{6385773ATERMSE_tumrgbd}. $*$ indicates methods relying on pre-trained data-driven priors.}
\vskip 0.1in
  \centering
  \resizebox{0.8\linewidth}{!}{
\begin{tabular}{lcccc}
\toprule
Method & \texttt{fr1/desk} & \texttt{fr2/xyz} & \texttt{fr3/office} & Avg. \\
\midrule
\rowcolor{gray!20}
\multicolumn{5}{l}{\textit{Neural Implicit Fields}} \\
NICE-SLAM~\cite{Zhu2021NICESLAM} & 4.3 & 31.7 & 3.9 & 13.3 \\
Vox-Fusion~\cite{Yang_Li_Zhai_Ming_Liu_Zhang_2022_voxfusion} & \textbf{3.5} & 1.5 & 26.0 & 10.3 \\
Point-SLAM~\cite{Sandström2023ICCVpointslam} & 4.3 & \textbf{1.3} & \textbf{3.5} & \textbf{3.0} \\
\hdashline
Loopy-SLAM$*$~\cite{liso2024loopyslam} & 3.8 & 1.6 & 3.4 & 2.9 \\
\midrule
\midrule
\rowcolor{gray!20}
\multicolumn{5}{l}{\textit{3D Gaussian Splatting}} \\
SplaTAM~\cite{keetha2024splatam} & 3.4 & 1.2 & 5.2 & 3.3 \\ 
GS-SLAM~\cite{yan2023gs} & 3.3 & 1.3 & 6.6 & 3.7 \\
Gaussian-SLAM~\cite{yugay2023gaussianslam} & 2.6 & 1.3 & 4.6 & 2.9 \\
\hdashline
LoopSplat$*$~\cite{zhu2024_loopsplat} & 2.1 & 1.6 & 3.2 & 2.3 \\
CG-SLAM$*$~\cite{hu2024cg} & 2.4 & 1.2 & 2.5 & 2.0 \\
\midrule
\midrule
Ours & \textbf{2.4} & \textbf{1.1} & \textbf{4.4} & \textbf{2.6} \\
\bottomrule
\end{tabular}}
\label{Tab:suppcamtumperscene}
\end{table*}

\begin{table*}
\caption{Rendering performance comparison in PSNR $\uparrow$, SSIM $\uparrow$, and LPIPS $\downarrow$ on TUM-RGBD~\cite{6385773ATERMSE_tumrgbd}. $*$ indicates methods relying on pre-trained data-driven priors.}
\vskip 0.1in
  \centering
  \resizebox{0.8\linewidth}{!}{
\begin{tabular}{llcccc}
\toprule
Method & Metric & \texttt{fr1/desk} & \texttt{fr2/xyz} & \texttt{fr3/office} & Avg. \\
\midrule
\rowcolor{gray!20}
\multicolumn{6}{l}{\textit{Neural Implicit Fields}} \\
\multirow{3}{*}{NICE-SLAM~\cite{Zhu2021NICESLAM}} 
& PSNR$\uparrow$  & 13.83 & \textbf{17.87} & 12.89 & 14.86 \\
& SSIM$\uparrow$ & 0.569 & \textbf{0.718} & 0.554 & 0.614 \\
& LPIPS$\downarrow$ & 0.482 & \textbf{0.344} & 0.498 & \textbf{0.441} \\
\midrule
\multirow{3}{*}{Vox-Fusion~\cite{Yang_Li_Zhai_Ming_Liu_Zhang_2022_voxfusion}} 
& PSNR$\uparrow$ & \textbf{15.79} & 16.32 & 17.27 & 16.46 \\
& SSIM$\uparrow$ & 0.647 & 0.706 & 0.677 & 0.677 \\
& LPIPS$\downarrow$ & 0.523 & 0.433 & 0.456 & 0.471 \\
\midrule
\multirow{3}{*}{ESLAM~\cite{johari-et-al-2023-ESLAM}} 
& PSNR$\uparrow$ & 11.29 & 17.46 & 17.02 & 15.26 \\
& SSIM$\uparrow$ & \textbf{0.666} & 0.310 & 0.457 & 0.478 \\
& LPIPS$\downarrow$ & \textbf{0.358} & 0.698 & 0.652 & 0.569 \\
\midrule
\multirow{3}{*}{Point-SLAM~\cite{Sandström2023ICCVpointslam}} 
& PSNR$\uparrow$ & 13.87 & 17.56 & \textbf{18.43} & \textbf{16.62} \\
& SSIM$\uparrow$ & 0.627 & 0.708 & \textbf{0.754} & \textbf{0.696} \\
& LPIPS$\downarrow$ & 0.544 & 0.585 & \textbf{0.448} & 0.526 \\
\hdashline
\multirow{3}{*}{Loopy-SLAM$*$~\cite{liso2024loopyslam}} 
& PSNR$\uparrow$ & - & - & - & 12.94 \\
& SSIM$\uparrow$ & - & - & - & 0.489 \\
& LPIPS$\downarrow$ & - & - & - & 0.645 \\

\midrule
\midrule
\rowcolor{gray!20}
\multicolumn{6}{l}{\textit{3D Gaussian Splatting}} \\
\multirow{3}{*}{SplaTAM~\cite{keetha2024splatam}} 
& PSNR$\uparrow$ & 22.00 & 24.50 & 21.90 & 22.80 \\
& SSIM$\uparrow$ & 0.857 & 0.947 & 0.876 & 0.893 \\
& LPIPS$\downarrow$ & 0.232 & 0.100 & 0.202 & 0.178 \\
\midrule
\multirow{3}{*}{Gaussian-SLAM~\cite{yugay2023gaussianslam}} 
& PSNR$\uparrow$  & 24.01 & 25.02 & 26.13 & 25.05 \\
& SSIM$\uparrow$ & 0.924 & 0.924 & 0.939 & 0.929 \\
& LPIPS$\downarrow$  & 0.178 & 0.186 & 0.141 & 0.168 \\
\hdashline
\multirow{3}{*}{LoopSplat$*$~\cite{zhu2024_loopsplat}} 
& PSNR$\uparrow$ & 22.03 & 22.68 & 23.47 & 22.72 \\
& SSIM$\uparrow$ & 0.849 & 0.892 & 0.879 & 0.873 \\
& LPIPS$\downarrow$ & 0.307 & 0.217 & 0.253 & 0.259 \\
\midrule
\midrule
\multirow{3}{*}{Ours} 
& PSNR$\uparrow$ & \textbf{27.09} & \textbf{33.01} & \textbf{30.50} & \textbf{30.20} \\
& SSIM$\uparrow$ & \textbf{0.959} & \textbf{0.982} & \textbf{0.974} & \textbf{0.972} \\
& LPIPS$\downarrow$ & \textbf{0.085} & \textbf{0.038} & \textbf{0.063} & \textbf{0.062} \\

\bottomrule
\end{tabular}}
\label{Tab:supprendertumperscene}
\end{table*}

\begin{table*}
\caption{Tracking performance comparisons in ATE RMSE $\downarrow [\mathrm{cm}]$ on ScanNet~\cite{scannet}. $*$ indicates methods relying on pre-trained data-driven priors.}
\vskip 0.1in
\label{Tab:suppcamscannetperscene}
  \centering
  \resizebox{0.8\linewidth}{!}{
\begin{tabular}{lccccccc}
\toprule
Method & \texttt{0000} & \texttt{0059} & \texttt{0106} & \texttt{0169} & \texttt{0181} & \texttt{0207} & Avg. \\
\midrule
\rowcolor{gray!20}
\multicolumn{8}{l}{\textit{Neural Implicit Fields}} \\
NICE-SLAM~\cite{Zhu2021NICESLAM} & 12.0 & 14.0 & \textbf{7.9} & \textbf{10.9} & \textbf{13.4} & \textbf{6.2} & \textbf{10.7} \\
Vox-Fusion~\cite{Yang_Li_Zhai_Ming_Liu_Zhang_2022_voxfusion} & 68.8 & 24.2 & 8.4 & 27.3 & 23.3 & 9.4 & 26.9 \\
Point-SLAM~\cite{Sandström2023ICCVpointslam} & \textbf{10.2} & \textbf{7.8} & 8.7 & 22.2 & 14.8 & 9.5 & 12.2 \\
\hdashline
Loopy-SLAM$*$~\cite{liso2024loopyslam} & 4.2 & 7.5 & 8.3 & 7.5 & 10.6 & 7.9 & 7.7 \\
\midrule
\midrule
\rowcolor{gray!20}
\multicolumn{8}{l}{\textit{3D Gaussian Splatting}} \\
SplaTAM~\cite{keetha2024splatam} & \textbf{12.8} & 10.1 & 17.7 & 12.1 & 11.1 & \textbf{7.5} & 11.9  \\ 
Gaussian-SLAM~\cite{yugay2023gaussianslam} & 24.8 & \textbf{8.6} & \textbf{11.3} & 14.6 & 18.7 & 14.4 & 15.4 \\
\hdashline
LoopSplat$*$~\cite{zhu2024_loopsplat} & 6.2 & 7.1 & 7.4 & 10.6 & 8.5 & 6.6 & 7.7 \\
CG-SLAM$*$~\cite{hu2024cg} & 7.1 & 7.5 & 8.9 & 8.2 & 11.6 & 5.3 & 8.1 \\
\midrule
\midrule
Ours & 17.8 & 8.7 & 11.8 & \textbf{10.5} & \textbf{10.6} & 8.6 & \textbf{11.3} \\

\bottomrule
\end{tabular}}
\end{table*}

\begin{table*}
\caption{Rendering performance comparison in PSNR $\uparrow$, SSIM $\uparrow$, and LPIPS $\downarrow$ on ScanNet~\cite{scannet}. $*$ indicates methods relying on pre-trained data-driven priors.}
\vskip 0.1in
  \centering
  \resizebox{0.8\linewidth}{!}{
\begin{tabular}{llccccccc}
\toprule
Method & Metric & \texttt{0000} & \texttt{0059} & \texttt{0106} & \texttt{0169} & \texttt{0181} & \texttt{0207} & Avg. \\
\midrule
\rowcolor{gray!20}
\multicolumn{9}{l}{\textit{Neural Implicit Fields}} \\
\multirow{3}{*}{NICE-SLAM~\cite{Zhu2021NICESLAM}} 
& PSNR$\uparrow$ & 18.71 & 16.55 & 17.29 & \textbf{18.75} & 15.56 & 18.38 & 17.54 \\
& SSIM$\uparrow$ & 0.641 & 0.605 & 0.646 & 0.629 & 0.562 & 0.646 & 0.621 \\
& LPIPS$\downarrow$ & 0.561 & 0.534 & 0.510 & 0.534 & 0.602 & 0.552 & 0.548 \\
\midrule
\multirow{3}{*}{Vox-Fusion~\cite{Yang_Li_Zhai_Ming_Liu_Zhang_2022_voxfusion}} 
& PSNR$\uparrow$ & 19.06 & 16.38 & \textbf{18.46} & 18.69 & 16.75 & 19.66 & 18.17 \\
& SSIM$\uparrow$ & 0.662 & 0.615 & \textbf{0.753} & 0.650 & 0.666 & 0.696 & 0.673 \\
& LPIPS$\downarrow$ & 0.515 & 0.528 & \textbf{0.439} & 0.513 & 0.532 & \textbf{0.500} & 0.504 \\
\midrule
\multirow{3}{*}{ESLAM~\cite{johari-et-al-2023-ESLAM}} 
& PSNR$\uparrow$ & 15.70 & 14.48 & 15.44 & 14.56 & 14.22 & 17.32 & 15.29 \\
& SSIM$\uparrow$ & 0.687 & 0.632 & 0.628 & 0.656 & 0.696 & 0.653 & 0.658 \\
& LPIPS$\downarrow$ & 0.449 & \textbf{0.450} & 0.529 & \textbf{0.486} & 0.482 & 0.534 & \textbf{0.488} \\
\midrule
\multirow{3}{*}{Point-SLAM~\cite{Sandström2023ICCVpointslam}} 
& PSNR$\uparrow$ & \textbf{21.30} & \textbf{19.48} & 16.80 & 18.53 & \textbf{22.27} & \textbf{20.56} & \textbf{19.82} \\
& SSIM$\uparrow$ & \textbf{0.806} & \textbf{0.765} & 0.676 & \textbf{0.686} & \textbf{0.823} & \textbf{0.750} & \textbf{0.751} \\
& LPIPS$\downarrow$ & \textbf{0.485} & 0.499 & 0.544 & 0.542 & \textbf{0.471} & 0.544 & 0.514 \\
\hdashline
\multirow{3}{*}{Loopy-SLAM$*$~\cite{liso2024loopyslam}} 
& PSNR$\uparrow$ & - & - & - & - & - & - & 15.23 \\
& SSIM$\uparrow$ & - & - & - & - & - & - & 0.629 \\
& LPIPS$\downarrow$ & - & - & - & - & - & - & 0.671 \\
\midrule
\midrule
\rowcolor{gray!20}
\multicolumn{9}{l}{\textit{3D Gaussian Splatting}} \\
\multirow{3}{*}{SplaTAM~\cite{keetha2024splatam}} 
& PSNR$\uparrow$ & 19.33 & 19.27 & 17.73 & 21.97 & 16.76 & 19.8 & 19.14 \\
& SSIM$\uparrow$ & 0.660 & 0.792 & 0.690 & 0.776 & 0.683 & 0.696 & 0.716 \\
& LPIPS$\downarrow$ & 0.438 & 0.289 & 0.376 & 0.281 & 0.420 & 0.341 & 0.358 \\
\midrule
\multirow{3}{*}{Gaussian-SLAM~\cite{yugay2023gaussianslam}} 
& PSNR$\uparrow$ & 28.54 & 26.21 & 26.26 & 28.60 & 27.79 & 28.63 & 27.70 \\
& SSIM$\uparrow$ & 0.926 & 0.934 & 0.926 & 0.917 & 0.922 & 0.914 & 0.923  \\
& LPIPS$\downarrow$ & 0.271 & 0.211 & 0.217 & 0.226 & 0.277 & 0.288 & 0.248 \\
\hdashline
\multirow{3}{*}{LoopSplat$*$~\cite{zhu2024_loopsplat}} 
& PSNR$\uparrow$ & 24.99 & 23.23 & 23.35 & 26.80 & 24.82 & 26.33 & 24.92 \\
& SSIM$\uparrow$ & 0.840 & 0.831 & 0.846 & 0.877 & 0.824 & 0.854 & 0.845 \\
& LPIPS$\downarrow$ & 0.450 & 0.400 & 0.409 & 0.346 & 0.514 & 0.430 & 0.425 \\
\midrule
\midrule
\multirow{3}{*}{Ours} 
& PSNR$\uparrow$ & \textbf{31.51} & \textbf{30.60} & \textbf{31.27} & \textbf{32.02} & \textbf{29.60} & \textbf{31.58} &  \textbf{31.10} \\
& SSIM$\uparrow$ & \textbf{0.957} & \textbf{0.974} & \textbf{0.975} & \textbf{0.962} & \textbf{0.954} & \textbf{0.946} & \textbf{0.961} \\
& LPIPS$\downarrow$ & \textbf{0.131} & \textbf{0.080} & \textbf{0.074} & \textbf{0.091} & \textbf{0.145} & \textbf{0.124} & \textbf{0.108} \\

\bottomrule
\end{tabular}}
\label{Tab:supprenderscannetperscene}
\end{table*}

\begin{table*}
\caption{Tracking performance comparisons in ATE RMSE $\downarrow [\mathrm{cm}]$ on ScanNet++~\cite{yeshwanthliu2023scannetpp}. $*$ indicates methods relying on pre-trained data-driven priors.}
\vskip 0.1in
  \centering
  \resizebox{0.8\linewidth}{!}{
\begin{tabular}{lcccccc}
\toprule
Method & \texttt{a} & \texttt{b} & \texttt{c} & \texttt{d} & \texttt{e} & Avg. \\
\midrule
\rowcolor{gray!20}
\multicolumn{7}{l}{\textit{Neural Implicit Fields}} \\
Point-SLAM~\cite{Sandström2023ICCVpointslam} & 246.16 & 632.99 & 830.79 & 271.42 & 574.86 & 511.24 \\
ESLAM~\cite{johari-et-al-2023-ESLAM} & \textbf{25.15} & \textbf{2.15} & \textbf{27.02} & \textbf{20.89} & \textbf{35.47} & \textbf{22.14} \\
\hdashline
Loopy-SLAM$*$~\cite{liso2024loopyslam} & - & - & 25.16 & 234.25 & 81.48 & 113.63 \\
\midrule
\midrule
\rowcolor{gray!20}
\multicolumn{7}{l}{\textit{3D Gaussian Splatting}} \\
SplaTAM~\cite{keetha2024splatam} & 1.50 & \textbf{0.57} & \textbf{0.31} & 443.10 & 1.58 & 89.41  \\ 
Gaussian-SLAM~\cite{yugay2023gaussianslam} & \textbf{1.37} & 5.97 & 2.70 & 2.35 & \textbf{1.02} & 2.68 \\
\hdashline
LoopSplat$*$~\cite{zhu2024_loopsplat} & 1.14 & 3.16 & 3.16 & 1.68 & 0.91 & 2.05 \\
\midrule
\midrule
Ours & 2.79 & 1.50 & 0.96 & \textbf{1.18} & 1.31 & \textbf{1.55}   \\

\bottomrule
\end{tabular}}
\label{Tab:suppcamscannetppperscene}
\end{table*}

\begin{table*}
\caption{Rendering performance comparison in PSNR $\uparrow$ on ScanNet++~\cite{yeshwanthliu2023scannetpp}. $*$ indicates methods relying on pre-trained data-driven priors.}
\vskip 0.1in
  \centering
  \resizebox{0.8\linewidth}{!}{
\begin{tabular}{lcccccc}
\toprule
Method & \texttt{a} & \texttt{b} & \texttt{c} & \texttt{d} & \texttt{e} & Avg. \\
\midrule
\rowcolor{gray!20}
\multicolumn{7}{l}{\textit{3D Gaussian Splatting}} \\
SplaTAM~\cite{keetha2024splatam} & 28.02 & 27.93 & 29.48 & 19.65 & 28.48 & 26.71 \\
Gaussian-SLAM~\cite{yugay2023gaussianslam} & 30.06 & 30.02 & 31.15 & 28.75 & 31.94 & 30.38  \\
\hdashline
LoopSplat$*$~\cite{zhu2024_loopsplat} & 30.15 & 30.08 & 30.04 & 28.94 & 31.78 & 30.20 \\
\midrule
Ours & \textbf{32.84} & \textbf{31.02} & \textbf{32.44} & \textbf{31.43} & \textbf{33.38} & \textbf{32.22}  \\

\bottomrule
\end{tabular}}
\label{Tab:supprenderscannetppperscene}
\end{table*}

\begin{table*}
\caption{Novel View Synsthesis performance comparison in PSNR $\uparrow$ on ScanNet++~\cite{yeshwanthliu2023scannetpp}. $*$ indicates methods relying on pre-trained data-driven priors.}
\vskip 0.1in
  \centering
  \resizebox{0.8\linewidth}{!}{
\begin{tabular}{lcccccc}
\toprule
Method & \texttt{a} & \texttt{b} & \texttt{c} & \texttt{d} & \texttt{e} & Avg. \\
\midrule
\rowcolor{gray!20}
\multicolumn{7}{l}{\textit{3D Gaussian Splatting}} \\
SplaTAM~\cite{keetha2024splatam} & 23.95 & 22.66 & 13.95 & 8.47 & 20.06 & 17.82 \\
Gaussian-SLAM~\cite{yugay2023gaussianslam} & \textbf{26.66} & \textbf{24.42} & 15.01 & 18.35 & 21.91 & 21.27 \\
\hdashline
LoopSplat$*$~\cite{zhu2024_loopsplat} & 25.60 & 23.65 & 15.87 & 18.86 & 22.51 & 21.30 \\
\midrule
Ours & 25.55 & 24.25 & \textbf{16.94} & \textbf{18.59} & \textbf{21.95} & \textbf{21.46} \\

\bottomrule
\end{tabular}}
\label{Tab:suppnvsrenderscannetppperscene}
\end{table*}

\begin{table*}
\caption{Rendering performance comparisons in PSNR $\uparrow$, SSIM $\uparrow$, and LPIPS $\downarrow$ on Replica~\cite{replica}. $*$ indicates methods relying on pre-trained data-driven priors.}
\vskip 0.1in
  \centering
  \resizebox{\linewidth}{!}{
\begin{tabular}{llccccccccc}
\toprule
Method & Metric & \texttt{Rm0} & \texttt{Rm1} & \texttt{Rm2} & \texttt{Off0} & \texttt{Off1} & \texttt{Off2} & \texttt{Off3} & \texttt{Off4} & Avg. \\
\midrule
\rowcolor{gray!20}
\multicolumn{11}{l}{\textit{Neural Implicit Fields}} \\
\multirow{3}{*}{NICE-SLAM~\cite{Zhu2021NICESLAM}} 
& PSNR$\uparrow$  & 22.12 & 22.47 & 24.52 & 29.07 & 30.34 & 19.66 & 22.23 & 24.94 & 24.42 \\
& SSIM$\uparrow$  & 0.689 & 0.757 & 0.814 & 0.874 & 0.886 & 0.797 & 0.801 & 0.856 & 0.809 \\
& LPIPS$\downarrow$  & 0.330 & 0.271 & 0.208 & 0.229 & 0.181 & 0.235 & 0.209 & 0.198 & 0.233 \\
\midrule
\multirow{3}{*}{Vox-Fusion~\cite{Yang_Li_Zhai_Ming_Liu_Zhang_2022_voxfusion}} 
& PSNR$\uparrow$  & 22.39 & 22.36 & 23.92 & 27.79 & 29.83 & 20.33 & 23.47 & 25.21 & 24.41 \\
& SSIM$\uparrow$  & 0.683 & 0.751 & 0.798 & 0.857 & 0.876 & 0.794 & 0.803 & 0.847 & 0.801 \\
& LPIPS$\downarrow$  & 0.303 & 0.269 & 0.234 & 0.241 & 0.184 & 0.243 & 0.213 & 0.199 & 0.236 \\
\midrule
\multirow{3}{*}{ESLAM~\cite{johari-et-al-2023-ESLAM}} 
& PSNR$\uparrow$  & 25.25 & 27.39 & 28.09 & 30.33 & 27.04 & 27.99 & 29.27 & 29.15 & 28.06 \\
& SSIM$\uparrow$  & 0.874 & 0.89 & 0.935 & 0.934 & 0.910 & 0.942 & 0.953 & 0.948 & 0.923 \\
& LPIPS$\downarrow$  & 0.315 & 0.296 & 0.245 & 0.213 & 0.254 & 0.238 & 0.186 & 0.210 & 0.245 \\
\midrule
\multirow{3}{*}{Point-SLAM~\cite{Sandström2023ICCVpointslam}} 
& PSNR$\uparrow$  & \textbf{32.40} & \textbf{34.08} & \textbf{35.50} & \textbf{38.26} & \textbf{39.16} & \textbf{33.99} & \textbf{33.48} & \textbf{33.49} & \textbf{35.17} \\
& SSIM$\uparrow$  & \textbf{0.974} & \textbf{0.977} & \textbf{0.982} & \textbf{0.983} & \textbf{0.986} & \textbf{0.960} & \textbf{0.960} & \textbf{0.979} & \textbf{0.975} \\
& LPIPS$\downarrow$  & \textbf{0.113} & \textbf{0.116} & \textbf{0.111} & \textbf{0.100} & \textbf{0.118} & \textbf{0.156} & \textbf{0.132} & \textbf{0.142} & \textbf{0.124} \\
\hdashline
\multirow{3}{*}{Loopy-SLAM$*$~\cite{liso2024loopyslam}} 
& PSNR$\uparrow$  & - & - & - & - & - & - & - & - & 35.47  \\
& SSIM$\uparrow$  & - & - & - & - & - & - & - & - & 0.981  \\
& LPIPS$\downarrow$  & - & - & - & - & - & - & - & - & 0.109 \\
\midrule
\midrule
\rowcolor{gray!20}
\multicolumn{11}{l}{\textit{3D Gaussian Splatting}} \\
\multirow{3}{*}{SplaTAM~\cite{keetha2024splatam}} 
& PSNR$\uparrow$  & 32.86 & 33.89 & 35.25 & 38.26 & 39.17 & 31.97 & 29.70 & 31.81 & 34.11 \\
& SSIM$\uparrow$  & 0.98 & 0.97 & 0.98 & 0.98 & 0.98 & 0.97 & 0.95 & 0.95 & 0.97 \\
& LPIPS$\downarrow$  & 0.07 & 0.10 & 0.08 & 0.09 & 0.09 & 0.10 & 0.12 & 0.15 & 0.10 \\
\midrule
\multirow{3}{*}{SGS-SLAM~\cite{li2024sgsslamsemanticgaussiansplatting}} 
& PSNR$\uparrow$  & 32.50 & 34.25 & 35.10 & 38.54 & 39.20 & 32.90 & 32.05 & 32.75 & 34.66 \\
& SSIM$\uparrow$  & 0.976 & 0.978 & 0.981 & 0.984 & 0.980 & 0.967 & 0.966 & 0.949 & 0.973 \\
& LPIPS$\downarrow$  & 0.070 & 0.094 & 0.070 & 0.086 & 0.087 & 0.101 & 0.115 & 0.148 & 0.096 \\
\midrule
\multirow{3}{*}{GS-SLAM~\cite{yan2023gs}} 
& PSNR$\uparrow$  & 31.56 & 32.86 & 32.59 & 38.70 & 41.17 & 32.36 & 32.03 & 32.92 & 34.27 \\
& SSIM$\uparrow$  & 0.968 & 0.973 & 0.971 & 0.986 & 0.993 & 0.978 & 0.970 & 0.968 & 0.975 \\
& LPIPS$\downarrow$  & 0.094 & 0.075 & 0.093 & 0.050 & 0.033 & 0.094 & 0.110 & 0.112 & 0.082 \\
\midrule
\multirow{3}{*}{MonoGS~\cite{MatsukiCVPR2024_monogs}} 
& PSNR$\uparrow$  & 34.83 & 36.43 & 37.49 & 39.95 & 42.09 & 36.24 & 36.70 & 36.07 & 37.50 \\
& SSIM$\uparrow$  & 0.954 & 0.959 & 0.965 & 0.971 & 0.977 & 0.964 & 0.963 & 0.957 & 0.960 \\
& LPIPS$\downarrow$  & 0.068 & 0.076 & 0.075 & 0.072 & 0.055 & 0.078 & 0.065 & 0.099 & 0.070 \\
\midrule
\multirow{3}{*}{Gaussian-SLAM~\cite{yugay2023gaussianslam}} 
& PSNR$\uparrow$ & 38.88 & 41.80 & 42.44 & 46.40 & 45.29 & 40.10 & 39.06 & 42.65 & 42.08 \\
& SSIM$\uparrow$  & \textbf{0.993} & \textbf{0.996} & \textbf{0.996} & \textbf{0.998} & \textbf{0.997} & \textbf{0.997} & \textbf{0.997} & \textbf{0.997} & \textbf{0.996} \\
& LPIPS$\downarrow$  & 0.017 & 0.018 & 0.019 & 0.015 & 0.016 & 0.020 & 0.020 & 0.020 & 0.018 \\
\hdashline
\multirow{3}{*}{LoopSplat$*$~\cite{zhu2024_loopsplat}} 
& PSNR$\uparrow$  & 33.07 & 35.32 & 36.16 & 40.82 & 40.21 & 34.67 & 35.67 & 37.10 & 36.63 \\
& SSIM$\uparrow$  & 0.973 & 0.978 & 0.985 & 0.992 & 0.990 & 0.985 & 0.990 & 0.989 & 0.985 \\
& LPIPS$\downarrow$  & 0.116 & 0.122 & 0.111 & 0.085 & 0.123 & 0.140 & 0.096 & 0.106 & 0.112 \\
\hdashline
\multirow{3}{*}{CG-SLAM$*$~\cite{hu2024cg}} 
& PSNR$\uparrow$  & 33.27 & - & - & - & - & - & 34.60 & - & - \\
& SSIM$\uparrow$  & - & - & - & - & - & - & - & - & - \\
& LPIPS$\downarrow$  & - & - & - & - & - & - & - & - & - \\
\midrule
\midrule
\multirow{3}{*}{Ours} 
& PSNR$\uparrow$  & \textbf{39.95} & \textbf{43.06} & \textbf{43.13} & \textbf{46.88} & \textbf{47.20} & \textbf{42.14} & \textbf{40.99} & \textbf{43.35} & \textbf{43.34} \\
& SSIM$\uparrow$  & 0.992 & \textbf{0.996} & \textbf{0.996} & \textbf{0.998} & \textbf{0.997} & 0.996 & 0.996 & 0.996 & \textbf{0.996} \\
& LPIPS$\downarrow$ & \textbf{0.014} & \textbf{0.013} & \textbf{0.014} & \textbf{0.009} & \textbf{0.009} & \textbf{0.012} & \textbf{0.013} & \textbf{0.015} & \textbf{0.012} \\

\bottomrule
\end{tabular}}
\label{Tab:supprenderreplicaperscene}
\end{table*}

\begin{table*}
\caption{Reconstruction performance comparison in Depth L1 [cm]$\downarrow$ and F1 $[\mathrm{\%}] \uparrow$ on Replica~\cite{replica}. $*$ indicates methods relying on pre-trained data-driven priors.}
\vskip 0.1in
  \centering
  \resizebox{\linewidth}{!}{
\begin{tabular}{llccccccccc}
\toprule
Method & Metric & \texttt{Rm0} & \texttt{Rm1} & \texttt{Rm2} & \texttt{Off0} & \texttt{Off1} & \texttt{Off2} & \texttt{Off3} & \texttt{Off4} & Avg. \\
\midrule
\rowcolor{gray!20}
\multicolumn{11}{l}{\textit{Neural Implicit Fields}} \\
\multirow{2}{*}{NICE-SLAM~\cite{Zhu2021NICESLAM}} 
& Depth L1 [cm]$\downarrow$ & 1.81 & 1.44 & 2.04 & 1.39 & 1.76 & 8.33 & 4.99 & 2.01 & 2.97 \\
& F1 $[\mathrm{\%}] \uparrow$ & 45.0 & 44.8 & 43.6 & 50.0 & 51.9 & 39.2 & 39.9 & 36.5 & 43.9 \\
\midrule
\multirow{2}{*}{Vox-Fusion~\cite{Yang_Li_Zhai_Ming_Liu_Zhang_2022_voxfusion}} 
& Depth L1 [cm]$\downarrow$ & 1.09 & 1.90 & 2.21 & 2.32 & 3.40 & 4.19 & 2.96 & 1.61 & 2.46 \\
& F1 $[\mathrm{\%}] \uparrow$ & 69.9 & 34.4 & 59.7 & 46.5 & 40.8 & 51.0 & 64.6 & 50.7 & 52.2 \\
\midrule
\multirow{2}{*}{ESLAM~\cite{johari-et-al-2023-ESLAM}} 
& Depth L1 [cm]$\downarrow$ & 0.97 & 1.07 & 1.28 & 0.86 & 1.26 & 1.71 & 1.43 & 1.06 & 1.18 \\
& F1 $[\mathrm{\%}] \uparrow$ & 81.0 & 82.2 & 83.9 & 78.4 & 75.5 & 77.1 & 75.5 & 79.1 & 79.1 \\
\midrule
\multirow{2}{*}{Co-SLAM} 
& Depth L1 [cm]$\downarrow$ & 0.99 & 0.82 & 2.28 & 1.24 & 1.61 & 7.70 & 4.65 & 1.43 & 2.59\\
& F1 $[\mathrm{\%}] \uparrow$ & 77.7 & 74.2 & 69.3 & 75.2 & 75.2 & 54.3 & 56.8 & 75.3 & 69.7 \\
\midrule
\multirow{2}{*}{Point-SLAM~\cite{Sandström2023ICCVpointslam}} 
& Depth L1 [cm]$\downarrow$ & \textbf{0.53} & \textbf{0.22} & \textbf{0.46} & \textbf{0.30} & \textbf{0.57} & \textbf{0.49} & \textbf{0.51} & \textbf{0.46} & \textbf{0.44} \\
& F1 $[\mathrm{\%}] \uparrow$ & \textbf{86.9} & \textbf{92.3} & \textbf{90.8} & \textbf{93.8} & \textbf{91.6} & \textbf{89.0} & \textbf{88.2} & \textbf{85.6} & \textbf{89.8} \\
\hdashline
\multirow{2}{*}{Loopy-SLAM$*$~\cite{liso2024loopyslam}} 
& Depth L1 [cm]$\downarrow$ & 0.30 & 0.20 & 0.42 & 0.23 & 0.46 & 0.60 & 0.37 & 0.24 & 0.35 \\
& F1 $[\mathrm{\%}] \uparrow$ & 91.6 & 92.4 & 90.6 & 93.9 & 91.6 & 88.5 & 89.0 & 88.7 & 90.8 \\
\midrule
\midrule
\rowcolor{gray!20}
\multicolumn{11}{l}{\textit{3D Gaussian Splatting}} \\
\multirow{2}{*}{SplaTAM~\cite{keetha2024splatam}} 
& Depth L1 [cm]$\downarrow$ & \textbf{0.43} & 0.38 & \textbf{0.54} & 0.44 & 0.66 & 1.05 & 1.60 & 0.68 & 0.72 \\
& F1 $[\mathrm{\%}] \uparrow$ & 89.3 & 88.2 & 88.0 & 91.7 & 90.0 & 85.1 & 77.1 & 80.1 & 86.1 \\
\midrule
\multirow{2}{*}{GS-SLAM~\cite{yan2023gs}} 
& Depth L1 [cm]$\downarrow$ & 1.31 & 0.82 & 1.26 & 0.81 & 0.96 & 1.41 & 1.53 & 1.08 & 1.16 \\
& F1 $[\mathrm{\%}] \uparrow$ & 62.9 & 79.9 & 66.8 & 80.0 & 81.6 & 66.0 & 59.2 & 65.0 & 70.2 \\
\midrule
\multirow{2}{*}{Gaussian-SLAM~\cite{yugay2023gaussianslam}} 
& Depth L1 [cm]$\downarrow$ & 0.61 & \textbf{0.25} & \textbf{0.54} & 0.50 & 0.52 & 0.98 & 1.63 & \textbf{0.42} & 0.68 \\
& F1 $[\mathrm{\%}] \uparrow$ & 88.8 & 91.4 & 90.5 & 91.7 & 90.1 & 87.3 & 84.2 & \textbf{87.4} & 88.9 \\
\hdashline
\multirow{2}{*}{LoopSplat$*$~\cite{zhu2024_loopsplat}} 
& Depth L1 [cm]$\downarrow$ & 0.39 & 0.23 & 0.52 & 0.32 & 0.51 & 0.63 & 1.09 & 0.40 & 0.51 \\
& F1 $[\mathrm{\%}] \uparrow$ & 90.6 & 91.9 & 91.1 & 93.3 & 90.4 & 88.9 & 88.7 & 88.3 & 90.4 \\
\midrule
\midrule
\multirow{2}{*}{Ours} 
& Depth L1 [cm]$\downarrow$ & 0.48 & 0.28 & 0.61 & \textbf{0.41} & \textbf{0.48} & \textbf{0.62} & \textbf{0.86} & 0.53 & \textbf{0.53} \\
& F1 $[\mathrm{\%}] \uparrow$ & \textbf{90.7} & \textbf{91.7} & \textbf{90.7} & \textbf{93.0} & \textbf{90.8} & \textbf{88.3} & \textbf{87.5} & 87.0 & \textbf{90.0} \\
\bottomrule

\end{tabular}}
\label{Tab:suppreconreplicaperscene}
\end{table*}

\begin{table*}
\caption{Tracking performance comparisons in ATE RMSE $\downarrow[\mathrm{m}]$ on KITTI~\cite{Geiger2012CVPR}.}
\vskip 0.1in
    \centering
\begin{tabular}{ccccc} 
\toprule
Sequence  & Gaussian-SLAM & SplaTAM & LoopSplat & Ours \\
\midrule
00 & 3.02 & 58.83 & 2.22 & \textbf{2.06} \\
\midrule
01 & 77.51 & 84.45 & 74.47 & \textbf{29.01} \\
\midrule
05 & 128.88 & 80.39 & 117.43 & \textbf{7.74} \\
\midrule
10 & 10.60 & 43.82 & 11.39 & \textbf{4.54} \\
\bottomrule
\end{tabular}
\label{Tab:suppkittitracking}
\end{table*}

\begin{table*}
\caption{Rendering performance comparisons in PSNR $\uparrow$ on KITTI~\cite{Geiger2012CVPR}.}
\vskip 0.1in
    \centering
\begin{tabular}{ccccc} 
\toprule
Sequence  & Gaussian-SLAM & SplaTAM & LoopSplat & Ours \\
\midrule
00 & 15.51	&9.82	&15.82	&\textbf{28.54} \\
\midrule
01 & 15.95	&12.89	&14.69	&\textbf{30.33} \\
\midrule
05 & 16.22	&26.48	&15.98	&\textbf{28.19} \\
\midrule
10 & 15.58	&25.58	&14.58	&\textbf{27.59} \\
\midrule
Peak GPU Use (GiB)	&2.74	&22.37	&3.56	&4.79 \\
\bottomrule
\end{tabular}
\label{Tab:suppkittimapping}
\end{table*}

\begin{table*}
\caption{Impact of depth noise and movability of Gaussians on the rendering performance in PSNR $\uparrow$, SSIM $\uparrow$, and LPIPS $\downarrow$ on Replica~\cite{replica}.}
\vskip 0.1in
\resizebox{\linewidth}{!}{
    \centering
\begin{tabular}{lccccc} 
\toprule
Metric & 10\% pixels w/ noises&	20\% pixels w/ noises&	30\% pixels w/ noises	&Gaussians movable along ray	&Ours(w/o additional noises \& fix)\\
\midrule
PSNR$\uparrow$	&43.41	&43.40	&43.29	&42.89	&43.06\\
\midrule
SSIM$\uparrow$	&0.996	&0.996	&0.996	&0.995	&0.996\\
\midrule
LPIPS$\downarrow$	&0.015	&0.015	&0.015	&0.020	&0.013\\
\bottomrule
\end{tabular}}
\label{Tab:suppdepthnoise}
\end{table*}


\end{document}